\documentclass{article}

\usepackage{geometry}

\newgeometry{vmargin={20mm}, hmargin={35mm,35mm}}   

\usepackage{microtype}
\usepackage{graphicx}
\usepackage{array}
\usepackage{booktabs} 
\usepackage{amsfonts}
\usepackage{amssymb,amsmath,amsthm}
\newtheorem{remark}{Remark}
\newtheorem{theorem}{Theorem}
\newtheorem{definition}{Definition}
\newtheorem{lemma}{Lemma}
\usepackage{soul}
\usepackage{lmodern}
\usepackage{booktabs}
\usepackage{threeparttable}
\usepackage{mathtools}

\usepackage{cuted}
\usepackage{color}

\usepackage{verbatim}

\usepackage[caption=false,subrefformat=parens,labelformat=parens]{subfig}

\DeclarePairedDelimiter\floor{\lfloor}{\rfloor}

\DeclareMathOperator{\spann}{span}

\usepackage{hyperref}

\def\gW{\mathcal{W}}

\newcommand*\di{\mathop{}\!\mathrm{d}}
\newcommand{\R}{\mathbb{R}}
\newcommand{\N}{\mathbb{N}}

\def\lf{\left\lfloor}   
\def\rf{\right\rfloor}

\title{A Deterministic Gradient-Based Approach to Avoid Saddle Points}

\author{
  Lisa Maria Kreusser \\
	Department of Applied Mathematics and Theoretical Physics\\
	University of Cambridge\\
	\texttt{L.M.Kreusser@damtp.cam.ac.uk}\\
   \and
  Stanley J. Osher \\
  Department of Mathematics\\
  University of California, Los Angeles\\
   \texttt{sjo@math.ucla.edu} \\
   \and
  Bao Wang \\
  Department of Mathematics\\
  Scientific Computing and Imaging Institute\\
  University of Utah\\
  \texttt{wangbaonj@gmail.com}
}

\begin{document}

\maketitle

\begin{abstract}
Loss functions with a large number of saddle points are one of the major obstacles for training modern machine learning models efficiently. First-order methods such as gradient descent are usually the methods of choice for training machine learning models. However, these methods converge to saddle points for certain choices of initial guesses. In this paper, we propose a modification of the recently proposed Laplacian smoothing gradient descent [Osher et al., arXiv:1806.06317], called modified Laplacian smoothing gradient descent (mLSGD), and demonstrate its potential to avoid saddle points without sacrificing the convergence rate. Our analysis is based on the attraction region, formed by all starting points for which the considered numerical scheme converges to a saddle point. We investigate the attraction region's dimension both analytically and numerically. For a canonical class of quadratic functions, we show that the dimension of the attraction region for mLSGD is $\floor*{(n-1)/2}$, and hence it is significantly smaller than that of the gradient descent whose dimension is $n-1$.
\end{abstract}

\section{Introduction}
Training machine learning (ML) models often reduces to solving the  {\em empirical risk minimization} problem \cite{vapnik1992principles} 
\begin{equation}
\label{eq:finite-sum}
\min_{\mathbf{x}\in\mathbb{R}^n} f(\mathbf{x}) 
\end{equation}
where $f\colon \R^n\to \R$ is the empirical risk functional, defined as
\begin{align*}
f(\mathbf{x}):=\frac{1}{N}\sum_{i=1}^N \mathcal{L}(\mathbf y_i,g(\mathbf{d}_i,\mathbf{x})).
\end{align*}
Here, the training set $\{(\mathbf{d}_i, \mathbf y_i)\}_{i=1}^N$ with $\mathbf d_i\in \R^n, \mathbf y_i\in \R^m$ for $n,m\in \N$ is given, 
$g\colon \R^n\times \R^n\to \R^m$ denotes 
the machine learning model parameterized by $\mathbf{x}$ 
and $\mathcal{L}(\mathbf y_i,g(\mathbf{d}_i, \mathbf{x}))$ is the training loss between the ground-truth label $\mathbf y_i\in \mathbb R^m$ and the model prediction  $g(\mathbf d_i, \mathbf{x})\in \mathbb R^m$. The training loss function $\mathcal L$ is typically a cross-entropy loss for classification and a root mean squared error for regression.  For many practical applications, $f$ is a highly nonconvex function, and $g$  is chosen among deep neural networks (DNNs), known for their remarkable performance across various applications. Deep neural network models are heavily overparametrized and require large amounts of training data. Both the number of samples $N$ and the dimension $n$ of $\mathbf{x}$ can scale up to millions or even billions \cite{he2016deep,russakovsky2015imagenet}. These complications pose serious computational challenges. Gradient descent (GD), stochastic gradient descent (SGD) and their momentum-accelerated variants are the method of choice for training high capacity ML models, since their merits include fast convergence, concurrence, and an easy implementation \cite{Bengio:2009,Rumelhart:1998,wang2020scheduled}. However, GD, or more generally first-order optimization algorithms relying only on gradient information, suffer from slow global convergence when saddle points exist \cite{Lee:2019}.

Saddle points are omnipresent in high-dimensional nonconvex optimization problems, and  correspond to highly suboptimal solutions of many machine learning models \cite{Dauphin:2014,Ge:2015,Huang-2015COLT,Sun:2018}. Avoiding the convergence to saddle points and the escape from saddle points are two interesting mathematical problems. The convergence to saddle points can be avoided by changing the dynamics of the optimization algorithms in such a way that their iterates are less likely or do not converge to saddle points. Escaping from saddle points ensures that iterates close to saddle points escape from them efficiently. Many methods have recently been proposed for the escape from saddle points. These methods are either based on adding noise to gradients  \cite{Du-2018NIPS,Jin-2017ICML,Jin-2018COLT,Levy:2016} or leveraging high-order information, such as Hessian, Hessian-vector product or relaxed Hessian information \cite{Agarwal:2017,Carmon:2019,Curtis:2019,Curtis:2014,Dauphin:2014,Liu:2017,Martens:2010,Nesterov:2006,Nocedal:2006,Paternain:2019}. To the best of our knowledge, little work has been done in terms of avoiding saddle points with only first-order information. 
 
GD is guaranteed to converge to first-order stationary points. However, it may get stuck at saddle points since only gradient information is leveraged. We call the region containing all starting points  from which the gradient-based algorithm converges to a saddle point the attraction region. While it is known that the attraction region associated with any strict saddle points is of measure zero \cite{Lee:2019,Lee:2016} under GD for sufficiently small step sizes, it is still one of the major obstacles for GD to achieve fast global  convergence; in particular, when there exist exponentially many saddle points \cite{ge2016:blog}. This work aims to avoid saddle points by reducing the dimension of the attraction region and is motivated by the Laplacian smoothing gradient descent (LSGD) \cite{LSGD:2018}.

\subsection{Our Contribution}

We propose the first deterministic first-order algorithm for avoiding saddle points where no noisy gradients or any high-order information is required. We quantify the efficacy of the proposed new algorithm in avoiding saddle points for a class of canonical quadratic functions and extend the results to general quadratic functions. We summarize our major contributions below.

\paragraph{A small modification of LSGD}

For solving minimization problems of the form \eqref{eq:finite-sum}, GD with  initial guess $\mathbf{x}^0\in \mathbb{R}^n$ can be applied and resulting in the following GD iterates  
$$
\mathbf{x}^{k+1}=\mathbf{x}^k-\eta \nabla f(\mathbf{x}^k),
$$
where $\eta>0$ denotes the step size. LSGD pre-multiplies the gradient by a Laplacian smoothing matrix with periodic boundary condition and leads to the following iterates
$$
\mathbf{x}^{k+1} = \mathbf{x}^k - \eta \left(\mathbf{I} - \sigma \mathbf{L}\right)^{-1}\nabla f(\mathbf{x}^k),
$$
where $\mathbf{I}$ is the $n\times n$ identity matrix and $\mathbf{L}$ is the discrete one-dimensional Laplacian, defined below in \eqref{eq:tri-diag}. LSGD can achieve significant improvements in training ML models \cite{10.1007/978-981-15-5232-8_47,LSGD:2018,10.1007/978-981-15-5232-8_14}, ML with differential privacy guarantees \cite{wang2020dp}, federated learning \cite{liang2020exploring}, and Markov Chain Monte Carlo sampling \cite{wang2019laplacian}. 

In this work, we propose a small modification of LSGD to  avoid saddle points efficiently. At its core is the replacement of the constant $\sigma$ in LSGD by an iteration-dependent function $\sigma(k)$, resulting in the  modified LSGD (mLSGD)
\begin{align}\label{eq:mlsgd}
\mathbf{x}^{k+1} = \mathbf{x}^k - \eta \left(\mathbf{I} - \sigma(k) \mathbf{L}\right)^{-1}\nabla f(\mathbf{x}^k).
\end{align}
For the analysis, we assume that $\sigma(k)$ is a non-constant, monotonic function that is constant at some point. 
With such a small modification on LSGD, we show that mLSGD has the same convergence rate as GD and can avoid saddle points efficiently.

\paragraph{Quantifying the avoidance of saddle points}
It is well-known that stochastic first-order methods like SGD rarely get stuck in saddle points, while standard first-order methods like GD may converge to saddle points. We show that small modifications of standard gradient-based methods such as mLSGD in \eqref{eq:mlsgd} outperform GD in terms of saddle point avoidance due to its smaller attraction region. To quantify the set of initial data which leads to the convergence to saddle points, we investigate the dimension of the attraction region. Low-dimensional attraction regions are equivalent to high-dimensional subspaces of initial data which can avoid saddle points. Since many nonconvex optimization problems can locally be approximated by quadratic functions, we restrict ourselves to quadratic functions with saddle points in the following which also reduces additional technical difficulties arising with general functions. We consider the class of quadratic functions   $f\colon \R^n\to\R$ with $f(\mathbf x)=\frac{1}{2}\mathbf x^T \mathbf B \mathbf x$  where we assume that $\mathbf B$ has both positive and negative eigenvalues to guarantee the existence of saddle points. For different matrices $\mathbf B$, our numerical experiments indicate that the dimension of the attraction region for the modified LSGD is a significantly smaller space than that for GD.

\paragraph{Analysing the dimension of the attraction region} 
For our analytical investigation of the avoidance of saddle points, we consider a canonical class of quadratic functions first, given by $f\colon \R^n\to \R$ with
\begin{equation}\label{objective}
f(x_1, \cdots, x_n) = \frac{c}{2}\left(\sum_{i=1}^{n-1} x_i^2 - x_n^2\right)
\end{equation}
with $c>0$. The attraction regions of  GD and the modified LSGD are given by
$$\mathcal{W}_{\text{GD}} = \left\{\mathbf{x}_0 \in \R^n \colon \mathbf{x}^{k+1} = \mathbf{x}^{k} -\eta \nabla f(\mathbf{x}^k)\text{ with }\lim_{k\to\infty} \mathbf{x}^k = \mathbf{0} \right\}$$
and
{\small
$$
\mathcal{W}_{\text{mLSGD}} = \left\{\mathbf{x}_0 \in \R^n \colon  \mathbf{x}^{k+1} = \mathbf{x}^{k} -\eta \left(\mathbf{I} - \sigma(k) \mathbf{L}\right)^{-1} \nabla f(\mathbf{x}^k)\text{ with }\lim_{k\to\infty} \mathbf{x}^k = \mathbf{0} \right\},
$$}
respectively, and are of dimensions
\begin{equation*}
{\rm dim} \mathcal{W}_{\text{GD}} = n-1,\qquad {\rm dim} \mathcal{W}_{\text{mLSGD}} = \floor*{\frac{n-1}{2}}.
\end{equation*}
These results indicate that the set of initial data converging to a saddle point is significantly smaller for the modified LSGD  than for GD. We extend these results to quadratic functions of the form $f(\mathbf x)=\frac{1}{2}\mathbf x^T \mathbf B \mathbf x$ where $\mathbf B\in \R^{n\times n}$ has both positive and  negative eigenvalues. In the two-dimensional case, the attraction region reduces to the trivial non-empty space $\{\mathbf 0\}$ for most choices of $\mathbf B$ unless a very peculiar condition on eigenvectors of $\mathbf B$ is satisfied, implying that saddle points can be avoided for any starting point of the iterative method \eqref{eq:mlsgd}.

\subsection{Notation}
We use boldface upper-case letters $\mathbf{A}$, $\mathbf{B}$ to denote matrices and boldface lower-case letters $\mathbf{x}$, $\mathbf{y}$ to denote vectors. The vector of zeros of length $n$ is denoted by $\mathbf 0\in \R^n$ and $A_{ij}$ denotes the entry $(i, j)$ of $\mathbf{A}$. For vectors we use $\|\cdot\|$ to denote the Euclidean norm, and for matrices we use $\|\cdot\|$ to denote spectral norm, respectively. 
The eigenvalues of $\mathbf A$ are denoted by $\lambda_i(\mathbf A)$ where we assume that they are ordered according to their real parts. For a function $f\colon  \mathbb{R}^n\to \mathbb{R}$, we use $\nabla f$  to denote its gradient.

\subsection{Organization}
This paper is structured as follows. In section \ref{algorithm-section}, we revisit the LSGD algorithm and motivate the modified LSGD algorithm. 
For quadratic functions with saddle points, we rigorously prove in section \ref{Avoid-Saddle} that the modified LSGD can significantly reduce the dimension of the attraction region. We provide a convergence analysis for the modified LSGD for nonconvex optimization in section \ref{Convergence-Rate}. Furthermore, in section \ref{Numerical-Results}, we  provide  numerical results  illustrating the avoidance of saddle points  of the modified LSGD in comparison to the standard GD. Finally, we conclude.

\section{Algorithm} \label{algorithm-section}

\subsection{LSGD}
Recently, Osher et al.\ \cite{LSGD:2018} proposed to replace the standard  or  stochastic gradient vector $\mathbf{y}\in \R^n$ by the  Laplacian smoothed surrogate $\mathbf{A}_\sigma^{-1} \mathbf{y}\in \R^n$
where 
\begin{equation}\label{eq:tri-diag}
\mathbf{A}_\sigma := \mathbf{I}-\sigma \mathbf{L}=
\begin{bmatrix}
1+2\sigma   & -\sigma &  0&\dots &0& -\sigma \\
-\sigma     & 1+2\sigma & -\sigma & \dots &0&0 \\
0 & -\sigma  & 1+2\sigma & \dots & 0 & 0 \\
\dots     & \dots & \dots &\dots & \dots & \dots\\
-\sigma     &0& 0 & \dots &-\sigma & 1+2\sigma
\end{bmatrix}
\end{equation}
for a positive constant $\sigma$, identity matrix $\mathbf I\in \R^{n\times n}$ and the discrete one-dimensional Laplacian $\mathbf L \in \R^{n\times n}$. The resulting numerical scheme reads
\begin{align}\label{LSGD}
\mathbf{x}^{k+1} = \mathbf{x}^k - \eta \mathbf{A}_\sigma^{-1}\nabla f(\mathbf{x}^k).
\end{align}
where GD is recovered for $\sigma=0$.
This simple Laplacian smoothing can help to avoid spurious minima, reduce the variance of SGD on-the-fly, and leads to better generalizations in training neural networks. Computationally, Laplacian smoothing can be implemented either by the Thomas algorithm together with the Sherman-Morrison formula in linear time, or by the fast Fourier transform (FFT) in quasi-linear time. For convenience, we use FFT to perform gradient smoothing where
$$
\mathbf{A}_\sigma^{-1} \mathbf{y} = {\rm ifft}\left(\frac{{\rm fft}(\mathbf{y})}{\mathbf{1} -\sigma \cdot {\rm fft}(\mathbf{d})}\right),
$$
with $\mathbf{d} = [-2, 1, 0, \cdots, 0, 1]^T\in \R^n$.

\subsection{Motivation for modifying LSGD to avoid saddle points}\label{sec:motivation}
To motivate the strength of modified LSGD methods in avoiding saddle points, we consider the two-dimensional setting, show the impact of varying $\sigma$ on the convergence to saddle points, and compare it to the convergence to saddle points for the standard LSGD with constant $\sigma$.
\subsubsection{Convergence to saddle points for LSGD}
For  given initial data $\mathbf x^0\in \R^2$, we apply LSGD \eqref{LSGD} for any constant $\sigma\geq 0$ to a quadratic function of the form $f(\mathbf x)=\frac{1}{2}\mathbf x^T \mathbf B \mathbf x$ where we suppose that $\mathbf{B}\in \R^{2\times 2}$ has one positive and one negative eigenvalue for the existence of a saddle point.
This yields
\begin{align}\label{eq:mlsgdexp}
    \mathbf{x}^{k+1} = \left( \mathbf I- \eta \mathbf{A}_\sigma^{-1}\mathbf B \right)\mathbf{x}^k=\left( \mathbf I- \eta \mathbf{A}_\sigma^{-1}\mathbf B \right)^{k+1}\mathbf{x}^0
\end{align}
where $$
\mathbf{A}_\sigma = \begin{bmatrix}
1+\sigma   & -\sigma \\
-\sigma     & 1+\sigma
\end{bmatrix}.$$
 Since $\mathbf{A}_\sigma^{-1}$ is positive definite, 
$\mathbf{A}_\sigma^{-1}\mathbf{B}$
has one positive and one negative eigenvalue, denoted by $\lambda_+$ and $\lambda_-$, respectively. We write $\mathbf{p}_+$ and $\mathbf{p}_-$ for the associated eigenvectors and we have $\mathbf x^0=\alpha_+ \mathbf p_++\alpha_- \mathbf p_-$ for scalars $\alpha_+,\alpha_-\in \R$. This implies
\begin{align*}
    \mathbf{x}^{k+1} =\alpha_+\left( 1- \eta \lambda_+ \right)^{k+1}\mathbf{p}_+ + \alpha_-\left( 1- \eta \lambda_- \right)^{k+1}\mathbf{p}_-.
\end{align*}
If $\mathbf x^0\in \spann\{\mathbf p_+\}$ or, equivalently, $\alpha_-=0$, we have
$\lim_{k\to \infty} \mathbf x^k=\mathbf 0$
for $\eta>0$ chosen sufficiently small such that  $|1- \eta \lambda_+|<1$ is satisfied. Hence, we have convergence to the unique saddle point in this case.

Alternatively, we can study the convergence to the saddle point by considering the ordinary differential equation 
associated with \eqref{LSGD}. For this, we investigate the limit $\eta\to 0$ and obtain
\begin{equation}
\label{Gradient-Flow}
\frac{\di \mathbf x}{\di t}= - \mathbf{A}_\sigma^{-1} 
\mathbf{B} \mathbf x
\end{equation}
with initial data $\mathbf x(0)=\mathbf x^0$. Since $\mathbf x^0=\alpha_+ \mathbf p_++\alpha_- \mathbf p_-$ for scalars $\alpha_+,\alpha_-\in \R$,
the solution to \eqref{Gradient-Flow} is given by
\begin{equation*}
\label{Gradient-Flow-Solution}
\mathbf x(t)= \alpha_+ \mathbf{p}_+\exp{(- \lambda_+ t)} + \alpha_- \mathbf{p}_-\exp{(- \lambda_- t)}.
\end{equation*}
If $\mathbf x^0\in \spann\{\mathbf p_+\}$,  the solution to \eqref{Gradient-Flow} reduces to 
\begin{equation*}
\mathbf x(t)= \alpha_+\mathbf{p}_+\exp{(- \lambda_+ t)}
\end{equation*}
and $\mathbf x(t)\to \mathbf 0$ as $t\to\infty$, i.e.,
LSGD for a constant $\sigma$ converges to the unique saddle point of $f$. 

This motivation can also be extended to the $n$-dimensional setting. For that, we consider $f(\mathbf x)=\frac{1}{2}\mathbf x^T \mathbf B \mathbf x$ where $\mathbf B\in \R^{n\times n}$ is a matrix with $k$ negative and $n-k$ positive eigenvalues. We assume that the eigenvalues are ordered, i.e.\  $\lambda_1\geq \ldots \geq \lambda_{n-k}>0> \lambda_{n-k+1}\geq \ldots\geq \lambda_n$, and the associated eigenvectors are denoted by $\mathbf p_1, \ldots, \mathbf p_n$. One can easily show that for any starting point in $\spann\{p_1,\ldots,p_{n-k}\}$, we have convergence to the saddle point, implying that the attraction region for GD or the standard LSGD is given by $\mathcal{W}_{\text{LSGD}}=\spann\{p_1,\ldots,p_{n-k}\}$ with $\dim \mathcal{W}_{\text{LSGD}}=n-k$.

\subsubsection{Avoidance of saddle points for the modified LSGD}\label{sec:motivationmlsgd}

\begin{figure}[!ht]
\centering
\includegraphics[width=\textwidth,trim={2cm 8.3cm 2cm 6.3cm},clip]{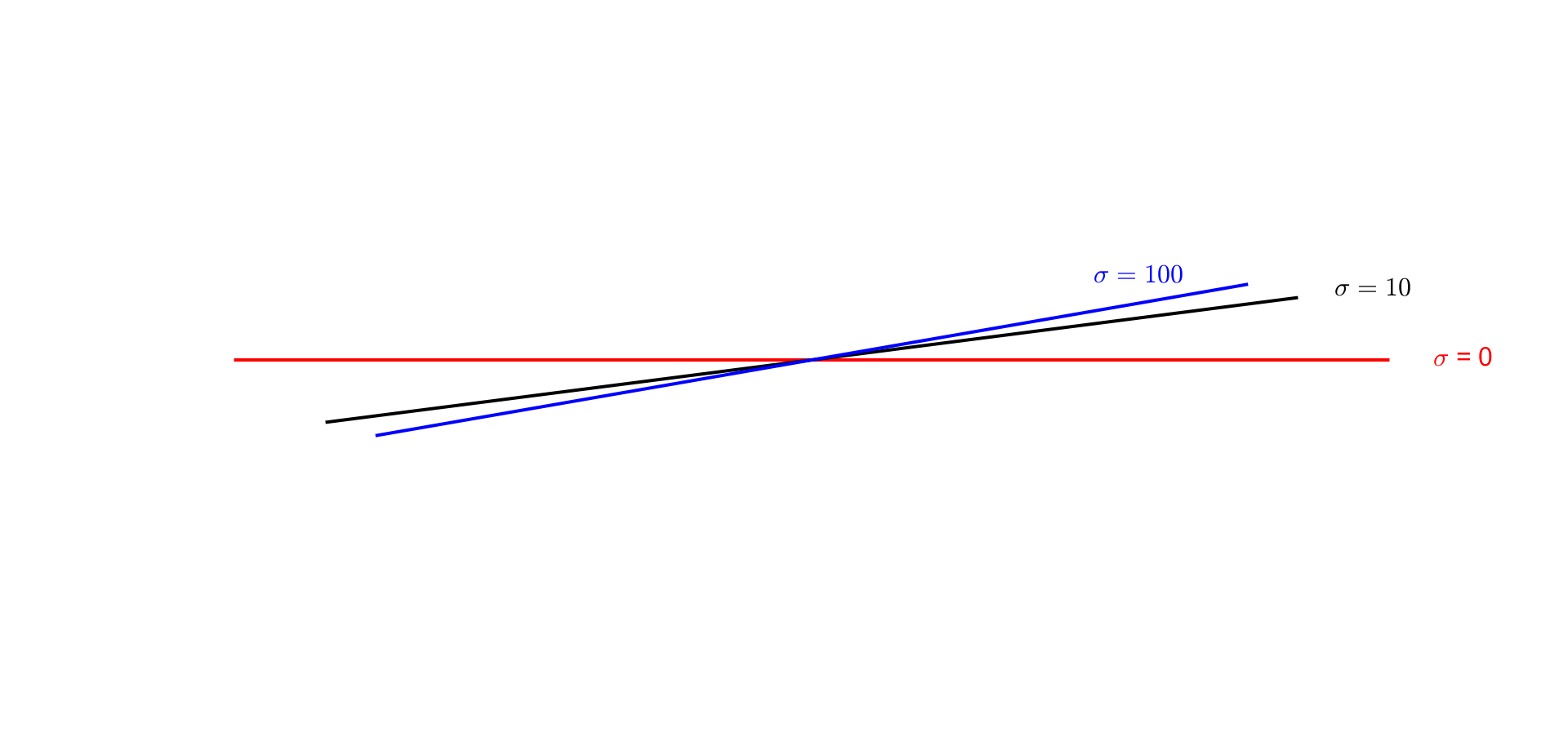}
\caption{The attraction region when  LSGD is applied to $f(\mathbf x)=\frac{1}{2}(x_1^2-x_2^2)$. The black, blue and red lines are the corresponding attraction regions for $\sigma=0$, 
$\sigma=10$ and $\sigma=100$, , respectively.
}
\label{fig:LSGD-Invariant-Manifold}
\end{figure}

In general, the eigenvectors and eigenvalues of $\mathbf A_\sigma^{-1} \mathbf B$ depend on $\sigma$. Hence, the behavior of the iterates $\mathbf x^k$ in \eqref{LSGD}  and their convergence to saddle points becomes more complicated for time-dependent functions $\sigma$ due to the additional time-dependence of eigenvectors and eigenvalues. 
To illustrate the impact of a time-dependent $\sigma$, we consider 
 the special case $f(\mathbf x)=\frac{1}{2}\mathbf x^T \mathbf B \mathbf x$ for
$$\mathbf B=\begin{bmatrix}
1   & 0 \\
0   & -1
\end{bmatrix},$$
i.e., $f(\mathbf x)=\frac{1}{2}(x_1^2-x_2^2)$ for $\mathbf x=[x_1,x_2]^T$.
The eigenvector $\mathbf{p}_+$ of $\mathbf A_\sigma^{-1}\mathbf B$, associated with the positive eigenvalue $\lambda_+$ of $\mathbf A_\sigma^{-1}\mathbf B$, is given by
$$
\mathbf{p}_+ = \begin{bmatrix}1\\ 0\end{bmatrix} \text{ for } \sigma=0 \quad \text{ and } \quad 
\mathbf{p}_+ = \frac{1}{\sqrt{\frac{(\sigma+1+\sqrt{2\sigma+1})^2+\sigma^2}{\sigma^2}}}\begin{bmatrix}\frac{\sigma+1+\sqrt{2\sigma+1}}{\sigma}\\ 1\end{bmatrix} \text{ for } \sigma>0.
$$
It is easy to see that $\nu(\sigma)=\frac{\sigma+1+\sqrt{2\sigma+1}}{\sigma}$ is a strictly decreasing function in $\sigma$ with $\nu\to+\infty$ as $\sigma\to 0$ and $\nu \to 1$ as $\sigma\to \infty$, implying that the corresponding normalised vector $\mathbf p_+$  is rotated counter-clockwise as $\sigma$ increases. Since $\mathbf p_+$ is given by $[\cos \phi,\sin \phi]^T$ for some $\phi\in [0,\frac{\pi}{4}]$, this implies that for   $\sigma_1,\sigma_2$ with $\sigma_1\neq \sigma_2$, the corresponding normalised eigenvectors $\mathbf{p}_+$ cannot be orthogonal and hence the associated normalised eigenvectors $\mathbf p_-$ cannot be orthogonal to each other. 
In particular, this is true for any bounded, strictly monotonic function $\sigma(k)$ of the iteration number $k$.
Figure~\ref{fig:LSGD-Invariant-Manifold} depicts the attraction regions for LSGD with constant values for $\sigma$, given by $\sigma=0$, $\sigma=10$ and $\sigma=100$, respectively.
Note that $\sigma=0$ corresponds to the standard gradient descent. Two  attraction regions intersect only at the origin, 
indicating  that starting from any point except $\mathbf 0$, LSGD   results in a slight change of direction in every time step  while $\sigma$ is strictly monotonic. This observation motivates that the modified LSGD for strictly monotonic $\sigma$ perturbs the gradient structure of $f$ in a non-uniform way while in the standard GD and LSGD the gradient is merely rescaled. In particular, the change of  direction of the iterates in every time step motivates the avoidance of the saddle point $\mathbf 0$ in the two-dimensional setting, while for LSGD with $\sigma$ constant, the iterates will converge to the saddle point for any starting point in $\spann\{\mathbf{p}_+\}$.

\subsection{Modified LSGD}
Based on the above heuristics,  we formulate the modified LSGD algorithm for 
positive, monotonic and bounded functions
$\sigma(k)$. The numerical scheme for the modified LSGD is given by
\begin{equation}
\label{LSGD-Vary-Sigma}
\mathbf{x}^{k+1} = \mathbf{x}^k - \eta \mathbf A_{\sigma(k)}^{-1}\nabla f(\mathbf{x}^k)
\end{equation}
where $\mathbf A_{\sigma(k)}=\mathbf{I} - \sigma(k)\mathbf{L}$.
The Laplacian smoothed surrogate
$\mathbf A_{\sigma(k)}^{-1}\nabla f(\mathbf{x}^k)$ 
can be computed by using either the Thomas algorithm or the FFT with  the same computational complexity as the standard LSGD.

\begin{remark}
For $\sigma(k)$ in the numerical scheme \eqref{LSGD-Vary-Sigma}, we choose a positive function which is easy to compute. Any 
positive, strictly monotonic and bounded function  $\sigma(k)$  guarantees the rotation of at least one eigenvector in the example in section \ref{sec:motivationmlsgd}.
\end{remark}

\section{Modified LSGD Can Avoid Saddle Points}\label{Avoid-Saddle}

In this section, we investigate the dimension of the attraction region for different classes of quadratic functions.

\subsection{Specific class of functions}
We consider the canonical class of quadratic functions in \eqref{objective} on $\R^n$ which has a unique saddle point at $\mathbf 0$. Note that this class of functions can be written as $f(\mathbf x)=\frac{c}{2}\mathbf x^T \mathbf B \mathbf x$ for some $c>0$,
where
\begin{equation*}
\mathbf{B} = 
\begin{bmatrix}
	1 & & &\\
	& \ddots & \\
	& & 1&\\
	& & & -1
\end{bmatrix}\in\R^{n\times n}. 
\end{equation*}
Since $c>0$ is a scaling factor which only influences the speed of convergence but not the direction of the updates, we can assume without loss of generality that $c=1$ in the following.
Starting from some point 
$\mathbf{x}^0\in\R^n$ and a given function $\sigma(k)$,
we apply the modified LSGD to $f$,  resulting in the  iterative scheme
\begin{equation}\label{Scheme}
\mathbf{x}^{k+1} = (\mathbf{I}-\eta \mathbf{A}_{\sigma(k)}^{-1}\mathbf{B})\mathbf{x}^k,
\end{equation}
where $\mathbf{A}_{\sigma(k)}$ is defined in \eqref{eq:tri-diag} for the function $\sigma=\sigma(k)$.

\begin{lemma}\label{prop:eighelp}
		For any $k\in\mathbb{N}$ fixed, the matrix $\mathbf{A}_{\sigma(k)}^{-1} \mathbf{B}$ is diagonalizable, its eigenvectors form a basis of $\mathbb{R}^n$  and the eigenvalues of the matrix $\mathbf{A}_{\sigma(k)}^{-1} \mathbf{B}$ satisfy 
		\begin{eqnarray*}
		1 \geq \lambda_1(\mathbf{A}_{\sigma(k)}^{-1} \mathbf{B}) \geq \ldots \geq \lambda_{n-1}(\mathbf{A}_{\sigma(k)}^{-1} \mathbf{B})
		> 0 >	\lambda_n(\mathbf{A}_{\sigma(k)}^{-1} \mathbf{B})\geq -1,
		\end{eqnarray*}
		where $\lambda_i(\mathbf{A}_{\sigma(k)}^{-1} \mathbf{B})$ denotes the $i$th largest eigenvalue of the matrix $\mathbf{A}_{\sigma(k)}^{-1} \mathbf{B}$.
		In particular, $\mathbf{A}_{\sigma(k)}^{-1} \mathbf{B}$ has exactly one negative eigenvalue.
\end{lemma}

\begin{proof}
        For ease of notation, we denote $\sigma(k)$ by $\sigma$ in the following.
		As a first step, we show that $\mathbf{A}_\sigma^{-1}\mathbf{B}$ is diagonalizable and its eigenvalues $\lambda_i(\mathbf{A}_\sigma^{-1}\mathbf{B})$  are real for $i=1,\ldots,n$. We prove this by showing that $\mathbf{A}_\sigma^{-1}\mathbf{B}$ is similar to a symmetric matrix. Note that $\mathbf{A}_\sigma^{-1}$ is a real, symmetric, positive definite matrix. Hence, $\mathbf{A}_\sigma^{-1}$ is diagonalizable with $\mathbf{A}_\sigma^{-1}=\mathbf{U} \mathbf{D} \mathbf{U}^T$ for an orthogonal matrix $\mathbf{U}$ and a diagonal matrix $\mathbf{D}$ with eigenvalues $\lambda_i(\mathbf{A}_\sigma^{-1})>0$ for $i=1,\ldots,n$ on the diagonal. This implies that there exists a real, symmetric, positive definite square root $\mathbf{A}_\sigma^{-1/2}=\mathbf{U} \sqrt{\mathbf{D}} \mathbf{U}^T$ with $\mathbf{A}_\sigma^{-1/2}\mathbf{A}_\sigma^{-1/2}=\mathbf{A}_\sigma^{-1}$ where $\sqrt{\mathbf{D}}$ denotes a diagonal matrix with diagonal entries $\sqrt{\lambda_i(\mathbf{A}_\sigma^{-1})}>0$. We have
		$$\mathbf{A}_\sigma^{1/2}\mathbf{A}_\sigma^{-1}\mathbf{B}\mathbf{A}_\sigma^{-1/2}=\mathbf{A}_\sigma^{-1/2}\mathbf{B}\mathbf{A}_\sigma^{-1/2}$$
		where $\mathbf{A}_\sigma^{-1/2}\mathbf{B}\mathbf{A}_\sigma^{-1/2}$ is symmetric due to the symmetry of $\mathbf{A}_\sigma^{-1/2}$ and $\mathbf{B}$. Thus, $\mathbf{A}_\sigma^{-1}\mathbf{B}$ is similar to the symmetric matrix $\mathbf{A}_\sigma^{-1/2}\mathbf{B}\mathbf{A}_\sigma^{-1/2}$. In particular, $\mathbf{A}_\sigma^{-1}\mathbf{B}$ is diagonalizable and has real eigenvalues like $\mathbf{A}_\sigma^{-1/2}\mathbf{B}\mathbf{A}_\sigma^{-1/2}$.
		
		Note that $\det(\mathbf{A}_\sigma^{-1}\mathbf{B})=\det(\mathbf{A}_\sigma^{-1}) \det(\mathbf{B})<0$ since $\det(\mathbf{A}_\sigma^{-1})>0$ and  $\det(\mathbf{B})=-1$. Since the determinant of a matrix is equal to the product of its eigenvalues and all eigenvalues of $\mathbf{A}_\sigma^{-1}\mathbf{B}$ are real, this implies that $\mathbf{A}_\sigma^{-1}\mathbf{B}$ has an odd number of negative eigenvalues. Next, we show that $\mathbf{A}_\sigma^{-1}\mathbf{B}$ has exactly one negative eigenvalue.  Defining
		$$\tilde{\mathbf{B}}:=	\begin{pmatrix}
		0 & & &\\
		& \ddots & \\
		& & 0&\\
		& & & -2
		\end{pmatrix}$$
		we have 
		$$\mathbf{A}_\sigma^{-1}\mathbf{B}=\mathbf{A}_\sigma^{-1}+\mathbf{A}_\sigma^{-1}\tilde{\mathbf{B}}$$
		where the matrix $\mathbf{A}_\sigma^{-1}\tilde{\mathbf{B}}$ has $n-1$-fold eigenvalue 0 and its last eigenvalue is given by $-2[\mathbf{A}_\sigma^{-1}]_{nn}$. We can write $\mathbf{A}_\sigma^{-1}=\frac{1}{\det \mathbf{A}_\sigma} \tilde{\mathbf{C}}$ where $\tilde{C}_{ij}=(-1)^{i+j}M_{ij}$ for the $(i,j)$-minor $M_{ij}$, defined as the determinant of the submatrix of $\mathbf{A}_\sigma$ by deleting the $i$th row and the $j$th column of $\mathbf{A}_\sigma$. Since all leading principal minors are positive for positive definite matrices, this implies that $M_{nn}>0$ due to the positive definiteness of $\mathbf{A}_\sigma$ and hence $[\mathbf{A}_\sigma]_{nn}>0$, implying that $\lambda_i(\mathbf{A}_\sigma^{-1}\tilde{\mathbf{B}})=0$ for $i=1,\ldots,n-1$ and $\lambda_n(\mathbf{A}_\sigma^{-1}\tilde{\mathbf{B}})<0$. The eigenvalues of $\mathbf{A}_\sigma^{-1}\mathbf{B}$ can now be estimated by Weyl's inequality for the sum of matrices, leading to  $$\lambda_{n-2}(\mathbf{A}_\sigma^{-1}\mathbf{B})\geq \lambda_{n-1}(\mathbf{A}_\sigma^{-1})+\lambda_{n-1}(\mathbf{A}_\sigma^{-1}\tilde{\mathbf{B}})>0$$ since the first term is positive and the second term is negative. Since $\mathbf{A}_\sigma^{-1}\mathbf{B}$ has an odd number of negative eigenvalues, this implies that $\lambda_{n-1}(\mathbf{A}_\sigma^{-1}\mathbf{B})>0$ and $\lambda_{n}(\mathbf{A}_\sigma^{-1}\mathbf{B})<0$. In particular, $\mathbf{A}_\sigma^{-1}\mathbf{B}$ has exactly one negative eigenvalue.
		
		To estimate upper and lower bounds of the eigenvalues of $\mathbf{A}_\sigma^{-1}\mathbf{B}$, note that the eigenvalues of the Laplacian $L$ are given by $2-2\cos(2\pi k/n)\in[0,4]$ for $k=0,\ldots,n/2$, implying that $\lambda_i(\mathbf{A}_\sigma)\in [1,1+4\sigma]$ and in particular, we have
		$$\lambda_i(\mathbf{A}_\sigma^{-1})\in \left[\frac{1}{1+4\sigma},1\right]$$
		for $i=1,\ldots,n$. Besides, we have 
		$$|\lambda_i(\mathbf{A}_\sigma^{-1}\mathbf{B})|\leq\rho(\mathbf{A}_\sigma^{-1}\mathbf{B})=\|\mathbf{A}_\sigma^{-1}\mathbf{B}\|\leq \|\mathbf{A}_\sigma^{-1}\|\|\mathbf{B}\|=\rho(\mathbf{A}_\sigma^{-1})\rho(\mathbf{B})\leq 1$$
		all $i=1,\ldots,n$, where $\rho(\mathbf{B})$ denotes the spectral radius of $\mathbf{B}$ and $\|\mathbf{B}\|$ denotes the operator norm of $\mathbf{B}$.
\end{proof}

Since $\mathbf{A}_{\sigma(k)}^{-1}\mathbf{B}$ is diagonalizable by Lemma \ref{prop:eighelp}, we can consider the invertible matrix $\mathbf{P}_{\sigma(k)}=(\mathbf{p}_{1, \sigma(k)},\ldots,\mathbf{p}_{n, \sigma(k)})$ whose columns $\mathbf{p}_{i,\sigma(k)}$ denote the normalized eigenvectors of $\mathbf{A}_{\sigma(k)}^{-1}\mathbf{B}$, associated with the eigenvalues $\lambda_i(\mathbf{A}_{\sigma(k)}^{-1}\mathbf{B})$, i.e.\ $$\lambda_i(\mathbf{A}_{\sigma(k)}^{-1}\mathbf{B})\mathbf{p}_{i,\sigma(k)}=\mathbf{A}_{\sigma(k)}^{-1}\mathbf{B} \mathbf{p}_{i,\sigma(k)}$$ for all $i=1,\ldots,n$, and $\{\mathbf{p}_{1, \sigma(k)},\ldots,\mathbf{p}_{n, \sigma(k)}\}$ forms a basis of unit vectors of $\R^n$. In the following,  $p_{i, j, \sigma(k)}$  denotes the $i$th entry of the $j$th eigenvector $\mathbf{p}_{j, \sigma(k)}$ of $\mathbf{A}_{\sigma(k)}^{-1}\mathbf{B}$, i.e.\ $\mathbf{p}_{j,\sigma(k)}=(p_{1, j,\sigma(k)}, \ldots, p_{n,j,\sigma(k)})$.

\begin{lemma}\label{prop:eigenvectors}
	 For any $k\in \N$ fixed, the matrix $\mathbf{A}_{\sigma(k)}^{-1}\mathbf{B}$ has $n-1$ eigenvectors $\mathbf{p}_{j,\sigma(k)}$
	 associated with positive eigenvalues. Of these, $\lfloor n/2\rfloor$ have the same form where the $l$th entry $p_{l,j,\sigma(k)}$ of the  $j$th eigenvector $\mathbf{p}_{j, \sigma(k)}$ satisfies \begin{align*} 
	 p_{l,j,\sigma(k)}\begin{cases} =p_{n-l,j,\sigma(k)}, & l=1,\ldots,n-1,\\
	  \neq 0, & l=n.\end{cases}
	 \end{align*}
	  For the remaining $\lfloor (n-1)/2\rfloor$ eigenvectors associated with positive eigenvalues,  the entry $p_{l,j,\sigma(k)}$ of  eigenvector $\mathbf{p}_{j,\sigma(k)}$ satisfies
		\begin{align}\label{eq:neigvalpos0}
	p_{l, j, \sigma(k)}=b\sin(l \theta_j),\quad l=1, \ldots, n,
	\end{align}
	where $\theta_j=\frac{2\pi m_j}{n}$ for some $m_j\in\mathbb{Z}$ and  $b \in\R\backslash \{0\}$ such that $\|\mathbf{p}_{j, \sigma(k)}\|=1$, implying  \begin{align*} 
	 p_{l,j,\sigma(k)}=\begin{cases} -p_{n-l,j,\sigma(k)}, & l=1,\ldots,n-1,\\
	  0, & l=n.\end{cases}
	 \end{align*}
	  The eigenvector $\mathbf{p}_{n,\sigma(k)}$ associated with the unique negative eigenvalue $\lambda_n(\mathbf{A}^{-1}_{\sigma(k)}\mathbf{B})$ satisfies \begin{align*} 
	 p_{l,n,\sigma(k)}\begin{cases} =p_{n-l,n,\sigma(k)}, & l=1,\ldots,n-1,\\
	 \neq 0, & l=n.\end{cases}
	 \end{align*}
\end{lemma}

\begin{proof}
    Since $k\in \N$ is fixed, we consider $\sigma$ instead of $\sigma(k)$ throughout the proof. 
    Besides, we  simplify the notation by dropping the index $\sigma=\sigma(k)$ in the notation of the eigenvectors $\mathbf{p}_{j,\sigma(k)}=(p_{1, j,\sigma(k)}, \ldots, p_{n,j,\sigma(k)})$, and we write $\mathbf{p}_j=(p_{1, j}, \ldots, p_{n,j})$ for $j=1,\ldots,n$.
    
	Since $\mathbf{A}_\sigma^{-1}\mathbf{B}$ and $\mathbf{B} \mathbf{A}_\sigma$ have the same eigenvectors and their eigenvalues are reciprocals, we can consider $\mathbf{B} \mathbf{A}_\sigma$ for determining the eigenvectors $\mathbf{p}_{j}$ for $j=1,\ldots,n$.	Note that the $n-1$ eigenvectors $\mathbf{p}_{j}$  of $\mathbf{B} \mathbf{A}_\sigma$ for $j=1,\ldots,n-1$ are associated with positive eigenvalues $\lambda_j(\mathbf{B} \mathbf{A}_\sigma)$ of $\mathbf{B} \mathbf{A}_\sigma$, while the eigenvector $\mathbf{p}_n$ is associated with the only negative eigenvalue  $\lambda_n(\mathbf{B} \mathbf{A}_\sigma)$. 
	 By introducing a slack variable $p_{0,j}$ we rewrite the eigenequation for the $j$th eigenvalue $\lambda_j(\mathbf{B}\mathbf{A}_\sigma)$, given by $$(\mathbf{B}\mathbf{A}_\sigma-\lambda_j(\mathbf{B}\mathbf{A}_\sigma))\mathbf{p}_j=\mathbf{0},$$ as
	\begin{align}
	\label{eq:ndim1}
	-\sigma p_{k-1,j}+(1+2\sigma -\lambda_j(\mathbf{B}\mathbf{A}))p_{k,j}-\sigma p_{k+1,j}=0, \quad
	k=1,\ldots,n-1,
	\end{align}
	with boundary conditions
	\begin{equation}
	\label{eq:ndim2}
	\sigma p_{1,j}+\sigma p_{n-1,j}-(1+2\sigma+\lambda_j(\mathbf{B}\mathbf{A}_\sigma))p_{n,j}=0
	\end{equation}
	and
	\begin{equation}
	\label{eq:ndim3}
	p_{0,j}=p_{n,j}.
	\end{equation}
	Equation \eqref{eq:ndim1} is a difference equation which can be solved by making the ansatz $p_{k,j}=r^k$.
	Plugging this ansatz into \eqref{eq:ndim1} results in the quadratic equation $$1-\frac{1+2\sigma-\lambda_j(\mathbf{B}\mathbf{A}_\sigma)}{\sigma}r+r^2=0$$ with solutions $r_{+/-}=d\pm \sqrt{d^2-1}$ where $$d:=\frac{1+2\sigma-\lambda_j(\mathbf{B}\mathbf{A}_\sigma)}{2\sigma}.$$
	Note that $r_+ r_-=d^2-(d^2-1)=1$ and $2d=r_++r_-=r_++(r_+)^{-1}$. 
	
	Let us consider the eigenvector $\mathbf{p}_n$ first. Since  $\lambda_n(\mathbf{B}\mathbf{A}_\sigma)<0$, this yields  $d>1$ and in particular $r_+\neq r_-$. We set $r:=r_+$, implying $r_-=1/r$, and obtain the general solution of the form $$p_{k,n}=b_1 r^k+b_2 r^{-k}, \quad k=0,\ldots,n$$ 
	for scalars $b_1,b_2\in\R$ which have to be determined from the boundary conditions \eqref{eq:ndim2},\eqref{eq:ndim3}. From \eqref{eq:ndim3} we obtain $$b_1+b_2=b_1 r^n+b_2 r^{-n},$$ implying that $b_1(1-r^n)=b_2 r^{-n}(1-r^n)$ and in particular $b_1=b_2 r^{-n}$ since  $r=r_+>1$. Hence, we obtain 
	\begin{equation}\label{eq:nansatzsym}
	p_{k,n}=b_1(r^k+r^{n-k}),\quad k=0,\ldots,n.
	\end{equation}
	For non-trivial solutions for the eigenvector $\mathbf{p}_n$ we require $b_1\neq 0$. Note that \eqref{eq:nansatzsym} implies that $p_{k,n}=p_{n-k,n}$ for $k=0,\ldots,n$. It follows from boundary condition \eqref{eq:ndim2} that $p_{n,n}\neq 0$ is necessary for non-trivial solutions.
	
	Next, we consider the $n-1$ eigenvectors $\mathbf{p}_{j}$  of $\mathbf{B} \mathbf{A}_\sigma$ associated with positive eigenvalues $\lambda_j(\mathbf{B} \mathbf{A}_\sigma)>0$ for $j=1,\ldots,n-1$.  Note that all positive eigenvalues of $\mathbf{B}\mathbf{A}_\sigma$ are in the interval  $[1,1+4\sigma]$ since  $\lambda_j(\mathbf{A}_\sigma)\in[1,1+4\sigma]$ and  $$\lambda_j(\mathbf{B}\mathbf{A}_\sigma)=\frac{1}{\lambda_j(\mathbf{A}_\sigma^{-1}\mathbf{B})}\geq 1$$
	 by Lemma \ref{prop:eighelp}. Hence, $\lambda_j(\mathbf{B}\mathbf{A}_\sigma)\leq \rho(\mathbf{B}\mathbf{A}_\sigma)=\|\mathbf{B}\mathbf{A}_\sigma\|\leq \|\mathbf{B}\|\|\mathbf{A}_\sigma\|=\rho(\mathbf{B})\rho(\mathbf{A}_\sigma)\leq 1+4\sigma$. Thus, it is sufficient to consider three different cases $\lambda_j(\mathbf{B}\mathbf{A}_\sigma)=1$, $\lambda_j(\mathbf{B}\mathbf{A}_\sigma)=1+4\sigma$ and $\lambda_j(\mathbf{B}\mathbf{A}_\sigma)\in(1,1+4\sigma)$. 
	 
	 We start by showing that all eigenvalues satisfy in fact $\lambda_j(\mathbf{B}\mathbf{A}_\sigma)\in(1,1+4\sigma)$. For this, assume that there exists $\lambda_j(\mathbf{B}\mathbf{A}_\sigma) =1$ for some $j\in\{1,\ldots,n-1\}$, implying that we have a single root	$r_+=r_-=d=1$. The general  solution to the difference equation \eqref{eq:ndim1} with boundary conditions \eqref{eq:ndim2}, \eqref{eq:ndim3} reads
	 $$p_{k,j}=(b_{1,j}+b_{2,j}k)r^k=b_{1,j}+b_{2,j}k, \quad k=0,\ldots,n$$
	 for constants $b_{1,j}, b_{2,j}\in\R$. Summing up all equations in \eqref{eq:ndim1} 
	 and subtracting \eqref{eq:ndim2} implies that $2p_{n,j}=0$, i.e.\ $p_{n,j}=0$. Hence,  \eqref{eq:ndim3} implies $p_{0,j}=p_{n,j}$ and our ansatz yields $0=p_{n,j}=p_{0,j}=b_{1,j}$. This results in $p_{k,j}=b_{2,j}k$ and $p_{n,j}=0=b_{2,j} n $ implies $b_{2,j}=0$. In particular, there exists no non-trivial solution and $\lambda_j(\mathbf{B}\mathbf{A}_\sigma)\neq 1$ for all $j=1,\ldots,n-1$. Next, we show that $\lambda_j(\mathbf{B}\mathbf{A}_\sigma)\neq 1+4\sigma$ for all $j=1,\ldots,n-1$ by contradiction. We assume that there exists $j\in\{1,\ldots,n-1\}$ such that $\lambda_j(\mathbf{B}\mathbf{A}_\sigma)= 1+4\sigma$, implying that $r_+=r_-=d=-1$. Due to the single root, the general solution is of the form 
	\begin{align*}
	p_{k,j}=(b_{1,j}+b_{2,j}k)r^k=(b_{1,j}+b_{2,j}k)(-1)^k,\quad k=0,\ldots,n.
	\end{align*}
	For $n$ even, \eqref{eq:ndim3} yields $b_{1,j}=p_{0,j}=p_{n,j}=b_{1,j}+b_{2,j}n$, implying $b_{2,j}=0$. Hence, the solution is constant with $p_{k,j}=b_{1,j}$, but does not satisfy boundary condition \eqref{eq:ndim2} unless $b_{1,j}=0$, resulting in the trivial solution. Similarly, we obtain for $n$ odd that $b_{1,j}=p_{0,j}=p_{n,j}=-b_{1,j}-b_{2,j}n$, implying $b_{2,j}=-2b_{1,j}/n$, i.e.\ $p_{k,j}=b_{1,j}(1-2k/n )(-1)^k$. Plugging this into the boundary condition \eqref{eq:ndim2} yields  $b_{1,j}=0$ since $\sigma>0$ and $n\geq 2$. In particular, there exists no non-trivial solution and the positive eigenvalues satisfy $\lambda_j(\mathbf{B}\mathbf{A}_\sigma)<1+4\sigma$ for all $j=1,\ldots,n-1$. Hence, we can now assume that $\lambda_j(\mathbf{B}\mathbf{A}_\sigma)\in(1,1+4\sigma)$. We conclude that $d\in(-1,1)$ and $r_{+/-}=d\pm i \sqrt{1-d^2}$ has two distinct roots. Setting $r:=r_+$ with $|r|=1$, we can introduce an angle $\theta$ and write $r=\exp(i\theta)=\cos \theta+i\sin\theta$, implying $d=\cos \theta$ and $r^k=\exp(ik\theta)$. Due to the distinct roots we consider the ansatz 
		$$p_{k,j}=b_{1,j}r^k+b_{2,j}r^{-k},\quad k=0,\ldots,n.$$
	The boundary condition \eqref{eq:ndim3}	implies $b_{1,j}r^n(1-r^n)=b_{2,j}(1-r^n)$ resulting in the two cases $r^n=1$ and $b_{1,j}r^n= b_{2,j}$. 
	
	For the  case $r^n=\cos(n\theta)+i\sin(n\theta)=1$, we  conclude that $\theta=2\pi m/n$ for some $m\in \mathbb{Z}$. This yields 
	\begin{align*}
	p_{k,j}=(b_{1,j}+b_{2,j})\cos(k \theta)+i(b_{1,j}-b_{2,j})\sin(k \theta),\\\quad k=0,\ldots,n,
	\end{align*}
	and we obtain 
	\begin{align*}
	p_{1,j}&=(b_{1,j}+b_{2,j})\cos(\theta)+i(b_{1,j}-b_{2,j})\sin(\theta), \\
	p_{n-1,j}&=(b_{1,j}+b_{2,j})\cos( \theta)-i(b_{1,j}-b_{2,j})\sin( \theta),\\
	p_{n,j}&=b_{1,j}+b_{2,j}.
	\end{align*}
	From boundary condition \eqref{eq:ndim2}, we obtain $$2\sigma(b_{1,j}+b_{2,j})\cos(\theta)-(1+2\sigma+\lambda_j(\mathbf{B}\mathbf{A}_\sigma))(b_{1,j}+b_{2,j})=0,$$
	implying that
	$b_{1,j}+b_{2,j}=0$ or $2\sigma\cos(\theta)=1+2\sigma+\lambda_j(\mathbf{B}\mathbf{A}_\sigma)$. Since $\lambda_j(\mathbf{B}\mathbf{A}_\sigma)>0$, the second case cannot be satisfied and we conclude $b_{1,j}+b_{2,j}=0$. This results in the general solution of the form 
	$p_{k,j}=2ib_{1,j}\sin(k \theta)$ for $k=0,\ldots,n$ for $b_{1,j}\in\mathbb{C}$, i.e., $\mathbf{p}_j=2ib_{1,j} (\sin(\theta),\ldots, \sin(n\theta))$. Rescaling by $1/(2i)$ results in the real eigenvectors $\mathbf{p}_j=(p_{1,j},\ldots,p_{n,j})$ whose entries are of the form \eqref{eq:neigvalpos0}
	where $b \in\R$ is chosen such that $\|\mathbf{p}_j\|=1$. Here, $p_{k,j}=-p_{n-k,j}$ for $k=1,\ldots,n-1$ and $p_{n,j}=0$. Further note that $p_{n/2,j}=0$ for $n$ even. By writing $\theta$ as $\theta_j=(2\pi m_j)/n$ for some $m_j\in \mathbb{Z}$, we can construct $(n-1)/2$ linearly independent eigenvectors for $n$ odd and $(n-2)/2$ for $n$ even, resulting in $\lfloor (n-1)/2 \rfloor$ linearly independent eigenvectors for any $n\in\N$. Since the matrix $\mathbf{A}_\sigma^{-1}\mathbf{B}$ is diagonalizable, there exist exactly $\lfloor (n-1)/2\rfloor$ normalized eigenvectors of the form \eqref{eq:neigvalpos0}.
	
	For $b_{1,j}r^n= b_{2,j}$, we obtain
	$$p_{k,j}=b_{1,j}(r^k+r^{n-k})=p_{n-k,j},\quad k=0,\ldots,n,$$
	i.e.\ the entries of $\mathbf{p}_j$ are arranged in the same way as the entries of $\mathbf{p}_n$. Further note that we can always set  $p_{n,j}\neq 0$, and additionally $p_{k,j}$ with $k=1,\ldots,n/2$ for $n$ even and $p_{k,j}$ with $k=1,\ldots,(n-1)/2$ for $n$ odd, resulting in a space of dimension $\lfloor n/2\rfloor+1$. Since $\mathbf{p}_1,\ldots,\mathbf{p}_n$ form a basis of $\R^n$, there are $\lfloor n/2\rfloor$ eigenvectors of this form, associated with positive eigenvalues.
\end{proof}

Lemma \ref{prop:eigenvectors} implies that  the matrix $\mathbf{A}_{\sigma(k)}^{-1}\mathbf{B}$ has one eigenvector $\mathbf{p}_{n, \sigma(k)}$ associated with the unique negative eigenvalue, $\lfloor (n-1)/2\rfloor$ eigenvectors of the form \eqref{eq:neigvalpos0} and $\lfloor n/2\rfloor$  eigenvalues associated with certain positive eigenvalues which are of the same form as $\mathbf{p}_{n, \sigma(k)}$. Note that $1+\lfloor (n-1)/2\rfloor+\lfloor n/2\rfloor=n$ for any $n\in\N$. 

In the following, we number the eigenvectors as follows. By $\mathbf{p}_{j,\sigma(k)}$ for $j=1, \ldots, \lfloor (n-1)/2\rfloor$ we denote the $\lfloor (n-1)/2\rfloor$ eigenvectors of the form \eqref{eq:neigvalpos0}. By $\mathbf{p}_{j,\sigma(k)}$ for $j=\lfloor (n-1)/2\rfloor+1, \ldots, n-1$, we denote the $\lfloor n/2 \rfloor$ eigenvectors of the form $p_{l, j, \sigma(k)}=p_{n-l,j,\sigma(k)}$ for $l=1,\ldots,n-1$ and $j=\lfloor (n-1)/2\rfloor+1, \ldots, n-1$. The eigenvectors $\mathbf{p}_{j,\sigma(k)}$ for $j=1,\ldots,n-1$, are associated  with positive eigenvalues, and $\mathbf{p}_{n,\sigma(k)}$ denotes the eigenvector associated with the unique negative eigenvalue. Similarly, we relabel the eigenvalues so that  eigenvalue $\lambda_j(\mathbf{A}_{\sigma(k)}^{-1}\mathbf{B})$ is associated with eigenvector $\mathbf{p}_{j,\sigma(k)}$. Using this basis of eigenvectors, we can write $\R^n=\mathcal{V} \oplus \mathcal{W}$ with
\begin{align}\label{eq:defspaces}
\begin{split}
	\mathcal{V}:&=\left\{ x_k=x_{n-k}, k=1,\ldots,\lf\frac{n-1}{2}\rf \right\},\\
	\gW:&=\left\{ x_k=-x_{n-k}, k=1,\ldots,\lf\frac{n-1}{2}\rf; x_n=0  \right\}.
	\end{split}
\end{align}
The spaces $\mathcal V,\mathcal W$ satisfy
\begin{align*}
\mathcal{V}=\spann \{\mathbf{p}_{\lfloor (n-1)/2\rfloor+1,\sigma(k)}, \ldots,\mathbf{p}_{n,\sigma(k)}\},\quad \mathcal{W}=\spann \{\mathbf{p}_{1,\sigma(k)},\ldots,\mathbf{p}_{\lfloor (n-1)/2\rfloor,\sigma(k)}\},
\end{align*}
where $\mathcal{V}$, $\mathcal{W}$ are orthogonal spaces and their definition is independent of $\sigma=\sigma(k)$ for any $k\in \N$.
For ease of notation, we introduce the set of indices $\mathcal{I}_\mathcal{V}:=\{\lfloor (n-1)/2\rfloor+1,\ldots,n\}$ and $\mathcal{I}_\mathcal{W}:=\{1,\ldots,\lfloor (n-1)/2\rfloor\}$ so that for all $k\in\N$ we obtain $\mathbf{p}_{i,\sigma(k)}\in\mathcal{V}$ for all $i\in\mathcal{I}_\mathcal{V}$ and $\mathbf{p}_{i,\sigma(k)}\in\mathcal{W}$ for all $i\in\mathcal{I}_\mathcal{W}$. In particular, $\mathbf{p}_{i,\sigma(k)}$ for $i\in\mathcal{I}_\mathcal{W}$ is independent of $\sigma$. 

\begin{remark}[Property of $\mathbf p_{n,\sigma(k)}$]\label{rem:orthonormpn}
It  follows immediately from the proof of Lemma \ref{prop:eigenvectors} that $\mathbf p_{n,\sigma}$ can be computed for any $\sigma\geq 0$. For $\sigma= 0$, we have $\mathbf  p_{n,0}=[0,\ldots,0,1]^T$ since $\mathbf A_0=\mathbf I$. For $\sigma>0$, the $k$th entry of $\mathbf p_{n,\sigma}$ is given by $p_{k,n,\sigma}=b_1(r^k+r^{n-k})$ for $k=1,\ldots,n$, by \eqref{eq:nansatzsym}, where the scalars $r>0$ and $b_1\in \R^n\backslash \{0\}$ depend on $\sigma$. Since all entries of $p_{n}$ are positive if $b_1>0$ and  negative if $b_1<0$, this implies that for  any $\sigma(k),\sigma(l)$ with $\sigma(k)\neq \sigma(l)$, we have $\mathbf p_{n,\sigma(k)}\cdot \mathbf p_{n,\sigma(l)}\neq 0$, i.e.\ the eigenvectors $\mathbf p_{n,\sigma(k)}, \mathbf p_{n,\sigma(l)}$ are not orthonormal to each other. Since any $\mathbf y\in\mathcal V$ can be written as a linear combination of $\mathbf p_{j,\sigma(k)}$ for $j\in \mathcal I_\mathcal V$, there exist $\beta_j$ for $j\in \mathcal I_\mathcal V$ with $\beta_n\neq 0$ such that $\mathbf{p}_{n,\sigma(l)}=\sum_{j\in \mathcal I_\mathcal V} \beta_j \mathbf{p}_{j,\sigma(k)}$.
\end{remark}

We have all the preliminary results to proof the main statement of this paper now:

\begin{theorem}\label{th:dimensionattraction}
Suppose that there exists $k_0>n-\lfloor (n-1)/2\rfloor$ such that $\sigma(k)=\sigma(k_0)$ for all $k\geq k_0$. For any $n\geq 2$ and $\mathbf{x}^0\notin \mathcal{W}\backslash\{0\}$, the modified LSGD scheme \eqref{Scheme} converges to the 
minimizer of $f$. The attraction region $\mathcal W$ satisfies \eqref{eq:defspaces} and is of dimension $\lfloor (n-1)/2\rfloor$.
\end{theorem}

	\begin{proof}
	Let $\mathbf{x}^0\notin \mathcal W\backslash\{0\}$ and let $k\in \N$ be given. We write $\mathbf{x}^k=\sum_{i=1}^n \alpha_{i,k}\mathbf{p}_{i,\sigma(k)}$  as $\mathbf{x}^k=\mathbf{w}^k+\mathbf{v}^k$ where 
	\begin{align*}
	\mathbf{v}^k:= \sum_{j\in \mathcal{I}_{\mathcal{V}}} \alpha_{j,k}\mathbf{p}_{j,\sigma(k)}\in\mathcal{V},\quad \mathbf{w}^k:= \sum_{j\in \mathcal{I}_{\mathcal{W}}} \alpha_{j,k}\mathbf{p}_{j,\sigma(k)}\in\mathcal{W}.
	\end{align*}
	Here, $\mathcal{V},\mathcal{W}$, defined in \eqref{eq:defspaces}, are independent of $\sigma$ with $\R^n=\mathcal{V}\oplus\mathcal{W}$. We apply the  modified   LSGD scheme \eqref{Scheme} and consider the sequence $\mathbf{x}^{k+1}=(\mathbf{I}- \eta \mathbf{A}_{\sigma(k)}^{-1}\mathbf{B}) \mathbf{x}^k=(\mathbf{I}- \eta \mathbf{A}_{\sigma(k)}^{-1}\mathbf{B}) \mathbf{w}^k+(\mathbf{I}-\eta \mathbf{A}_{\sigma(k)}^{-1}\mathbf{B}) \mathbf{v}^k$.
	We define $\mathbf{w}^{k+1}=(\mathbf{I}- \eta \mathbf{A}_{\sigma(k)}^{-1}\mathbf{B}) \mathbf{w}^k\in\mathcal{W}$ and $\mathbf{v}^{k+1}=(\mathbf{I}-  \eta \mathbf{A}_{\sigma(k)}^{-1}\mathbf{B}) \mathbf{v}^k\in\mathcal{V}$ iteratively. 
	Since $\mathbf{p}_{j,\sigma(k)}$ and the associated eigenvalues $\lambda_j(\mathbf{A}^{-1}_{\sigma(k)} \mathbf{B})$ are in fact independent of $\sigma(k)$ for $j\in\mathcal{I}_\mathcal{W}$, we have $\mathbf{p}_{j,\sigma(k+l)}=\mathbf{p}_{j,\sigma(k)}$ for any $l\geq 0$. We obtain
	\begin{align*}
	\mathbf{w}^{k+l} &=\left( \prod_{j=1}^l (\mathbf{I}- \eta \mathbf{A}_{\sigma(k+j)}^{-1}\mathbf{B})\right) \mathbf{w}^{k}= \sum_{j\in\mathcal{I}_\mathcal{W}} \alpha_{j,k} (1-\eta\lambda_j(\mathbf{A}_{\sigma(k)}^{-1}\mathbf{B}))^l \mathbf{p}_{j,\sigma(k)}
	\end{align*}
	 for any $l\geq 0$. By Lemma \ref{prop:eighelp}, the
	eigenvalues $\lambda_j(\mathbf{A}^{-1}_{\sigma(k)} \mathbf{B})$ satisfy $1-\eta\lambda_j(\mathbf{A}^{-1}_{\sigma(k)} \mathbf{B}) \in (0,1)$ for $j\in\mathcal{I}_\mathcal{W}$ and any $\eta\in(0,1)$, implying $\mathbf{w}^k\to 0$ as $k\to \infty$.
	For proving the unboundedness of $\mathbf{x}^k$  as $k\to\infty$ it is hence sufficient to show that $\mathbf{v}^k$ is unbounded as $k\to\infty$ for any $\mathbf{v}^0\in \mathcal V\backslash \{0\}$. 
	Since $\sigma=\sigma(k)$ is constant for all $k\geq k_0$, we have
	\begin{align*}
	\mathbf{v}^{k_0+l} &=\left( \prod_{j=1}^l (\mathbf{I}- \eta \mathbf{A}_{\sigma(k_0+j)}^{-1}\mathbf{B})\right) \mathbf{v}^{k_0}
	=\sum_{j\in\mathcal{I}_\mathcal{V}}\alpha_{j,k_0}(1-\eta\lambda_j(\mathbf{A}_{\sigma(k_0)}^{-1}\mathbf{B}))^l\mathbf{p}_{j,\sigma(k_0)}
	\end{align*}
	for any $l\geq 0$.
    Since $|1-\eta\lambda_j(\mathbf{A}_{\sigma(k_0)}^{-1}\mathbf{B})|<1$ for $\eta\in (0,1)$ and all $j=1,\ldots,n-1$, and $|1-\eta\lambda_n(\mathbf{A}_{\sigma(k_0)}^{-1}\mathbf{B})|>1$ by Lemma \ref{prop:eighelp},   $\mathbf{v}^{k_0+l}$ is unbounded as $l\to\infty$ if and only if $\alpha_{n,k_0}\neq 0$. 
    We show that starting from $\mathbf{v}^0\in \mathcal V\backslash \{0\}$ there exists $k\geq 0$ such that ${\mathbf{v}}^{k} =\sum_{i\in\mathcal{I}_\mathcal{V}} \alpha_{i,k} \mathbf{p}_{i,\sigma(k)} \in\mathcal{V}\backslash \{\mathbf{0}\}$ with $\alpha_{n,k}\neq 0$ and by Remark \ref{rem:orthonormpn} this guarantees $\alpha_{n,l}\neq 0$ for all $l\geq k$. 
	
	Starting from $\mathbf{v}^{0}=\sum_{i\in\mathcal{I}_\mathcal{V}} \alpha_{i,0} \mathbf{p}_{i,\sigma(0)}\neq \mathbf{0}$ we can assume that $\alpha_{n,0}=0$.
	Note that $\mathbf{v}^{0}$ is a function of $|\mathcal{I}_\mathcal{V}|=n-\lfloor (n-1)/2 \rfloor$ parameters where one of the parameter in the linear combination  can be regarded as a scaling parameter and thus, it can be set as any constant. This results in $n-\lfloor (n-1)/2 \rfloor-2 $  parameters which can be adjusted in such a way that $\mathbf{v}^{k}=\sum_{i\in\mathcal{I}_\mathcal{V}} \alpha_{i,k} \mathbf{p}_{i,\sigma(k)}$ with $\alpha_{n,k}=0$ for $k=0,\ldots,k_e$ with $k_e= n-\lfloor (n-1)/2 \rfloor-2$. We can determine these $n-\lfloor (n-1)/2 \rfloor-1$ parameters from $n-\lfloor (n-1)/2 \rfloor-1$ conditions, resulting in a linear system of $n-\lfloor (n-1)/2 \rfloor-1$ equations. However, 
	the additional condition
	\begin{align*}
	\mathbf{v}^{k_e+1} &=\Pi_{i=0}^{k_e}(\mathbf{I} -\eta\mathbf{A}_{\sigma(i)}^{-1}\mathbf{B})\mathbf{v}^{0}=\sum_{i\in\mathcal{I}_\mathcal{V}} \alpha_{i,k_e+1} \mathbf{p}_{i,\sigma(k_e)}
	\end{align*}
	with $\alpha_{n,k_e+1}=0$ leads to the unique trivial solution of the full linear system of size $n-\lfloor (n-1)/2\rfloor$, i.e., the assumption  $\mathbf{v}^{0}\neq \mathbf 0$ is not satisfied. This implies that for any $\mathbf{v}^{0}\in\mathcal{V}\backslash \{0\}$ a vector $\mathbf{v}^{k}=\sum_{i\in\mathcal{I}_\mathcal{V}} \alpha_{i,k}\mathbf{p}_{i,\sigma(k)}$ with $\alpha_{n,k}\neq 0$ is reached in finitely (after at most $n-\lfloor (n-1)/2\rfloor-1$) steps.
\end{proof}

To sum up, in Theorem \ref{th:dimensionattraction} we have discussed the convergence of the modified LSGD for the  canonical class of quadratic functions in \eqref{objective} on $\R^n$. We showed:
\begin{itemize}
    \item The attraction region  $\mathcal W$ of the modified LSGD is given by \eqref{eq:defspaces} with $\dim \mathcal W=\lfloor (n-1)/2\rfloor$.
    \item The definition of the attraction region $\mathcal W$ is given by the linear subspace of  eigenvectors of $\mathbf A_\sigma^{-1}\mathbf B$ which are independent of $\sigma$.
    \item The attraction region of the modified LSGD is significantly smaller than for the attraction region $\mathcal W_{LSGD}$ of the standard GD or the standard LSGD with $\dim \mathcal W_{LSGD}=n-1$.
    \item For any $\mathbf{x}^0\notin \mathcal{W}$, the modified LSGD scheme in \eqref{Scheme} converges to the minimizer.
    \item In the 2-dimensional setting, the attraction region of the modified LSGD  satisfies $\mathcal{W}=\{0\}$ and is of dimension zero. For any $\mathbf x^0\neq \mathbf 0$ the modified LSGD converges to the minimizer in this case.
    \item The proof of Theorem \ref{th:dimensionattraction} only considers the subspaces $\mathcal V,\mathcal W$ and uses the independence of $\sigma$ of the eigenvectors in $\mathcal{W}$. This observation is crucial for extending the results in Theorem~\ref{th:dimensionattraction} to any matrix $\mathbf B\in \R^{n\times n}$ with at least one positive and one negative eigenvalue.
\end{itemize}

\subsection{Extension to quadratic functions with saddle points}

While we investigated the convergence to saddle points for a  canonical class of quadratic functions in Theorem \ref{th:dimensionattraction}, we consider quadratic functions of the form $f(\mathbf x)=\frac{1}{2}\mathbf x^T \mathbf B \mathbf x$ for $\mathbf B\in \R^{n\times n}$. First, we suppose that the saddle points of $f$ are non-degenerate, i.e., all eigenvalues of $\mathbf B\in \R^{n\times n}$ are non-zero. For the existence of saddle points, we require that there exist at least one positive and one negative eigenvalue of $\mathbf B$.

Suppose that $\mathbf B$ has $k$ negative and $n-k$ positive eigenvalues. Since $\mathbf A_{\sigma}$ is positive definite for any $\sigma\geq 0$, all its eigenvalues are positive and hence $\mathbf A_\sigma^{-1} \mathbf B$ has $k$ negative and $n-k$ positive eigenvalues. Due to the conclusion from Theorem \ref{th:dimensionattraction}, it is sufficient to determine the space $\mathcal W$, consisting of all eigenvectors of $\mathbf A_\sigma^{-1}\mathbf B$ which are independent of $\sigma$ and are associated with positive eigenvalues.

Let $\sigma>0$ be given and suppose that  $\mathbf p\in \mathcal W\backslash \{\mathbf 0\}$. Then, $\mathbf p$ is an eigenvector of $\mathbf A_\sigma^{-1}\mathbf B$ and $\mathbf B$ corresponding to eigenvalues $\lambda(\mathbf A_\sigma^{-1} \mathbf B)>0$ and $\lambda(\mathbf B)>0$, respectively. By the definition of $\mathbf p$, we have
$$\lambda(\mathbf B) \mathbf p=\mathbf B \mathbf p=\lambda(\mathbf A_\sigma^{-1} \mathbf B) \mathbf A_\sigma\mathbf p=\lambda(\mathbf A_\sigma^{-1} \mathbf B) \mathbf p-\sigma \lambda(\mathbf A_\sigma^{-1} \mathbf B) \mathbf L \mathbf p$$
where we used the definition of $\mathbf A_\sigma$ in \eqref{eq:tri-diag}. We conclude that $\mathbf p\in\mathcal W$ if and only if $\mathbf L \mathbf p\in \spann\{ \mathbf p\}$.

\subsubsection{The 2-dimensional setting}\label{sec:extension2d}

For $n=2$, the eigenvectors of $\mathbf L$ are given by 
\begin{align*}
   \mathbf v_1=\frac{1}{\sqrt{2}}\begin{bmatrix}
     1\\ 1
    \end{bmatrix},\qquad  \mathbf v_2=\frac{1}{\sqrt{2}}\begin{bmatrix}
     1\\ -1
    \end{bmatrix}, 
\end{align*}
associated with the eigenvalues $0$ and $-4$, respectively. Since  $\mathbf p\in\mathcal W$ can be written as $\mathbf p=\alpha_1 \mathbf v_1+\alpha_2 \mathbf v_2$ for  coefficients $\alpha_1,\alpha_2\in \R$, we have $\mathbf L \mathbf p=-4\alpha_2 \mathbf v_2$. The condition $\mathbf L \mathbf p \in \spann\{p\}$ implies that $\mathbf p\in \spann\{\mathbf v_1\}$ or $\mathbf p\in \spann\{\mathbf v_2\}$.

We require $\mathbf B\in \R^{2\times 2}$ has one positive and one negative eigenvalue for the existence of saddle points, i.e.\ $\mathbf B$ is diagonalisable. We conclude that $\dim \mathcal W =1$ if and only if 
\begin{align*}
    \mathbf{B}=\begin{bmatrix}
     \mathbf v & \mathbf w
    \end{bmatrix} \begin{bmatrix}
      \mu_1 & 0 \\ 0 & \mu_2
    \end{bmatrix}\begin{bmatrix}
     \mathbf v & \mathbf w
    \end{bmatrix}^T
\end{align*}
where $\mu_1>0>\mu_2$ with $\mathbf v \in\spann\{\mathbf v_1\}$ or $\mathbf v \in\spann\{\mathbf v_2\}$. Examples of matrices with $\dim \mathcal W=1$ include
\begin{align*}
   \mathbf B_1 =\begin{bmatrix}
 0 & 1 \\ 1 & 0
\end{bmatrix}\quad \text{and}\quad \mathbf B_2 =\begin{bmatrix}
 0 & -1 \\ -1 & 0
\end{bmatrix}
\end{align*}
which correspond to the functions $f(\mathbf x)=x_1 x_2$ and $f(\mathbf x)=-x_1 x_2$ for   $\mathbf x=[x_1, x_2]^T$, respectively. Since the eigenvector associated with the positive eigenvalue  does not satisfy the above condition for most matrices $\mathbf B\in \R^{2\times 2}$, we have $\dim\mathcal W=0$ for most $2$-dimensional examples, including the canonical class discussed in Theorem \ref{th:dimensionattraction}.

\subsubsection{The $n$-dimensional setting}
Similar to the proof in Lemma \ref{prop:eighelp}, one can show that the eigenvalues of the positive semi-definite matrix $-\mathbf L\in \R^{n\times n}$ have a specific form. We denote the $n$ eigenvalues of $\mathbf L$ by $\lambda_1,\ldots,\lambda_n$ where $0=\lambda_1>\lambda_2\geq \ldots\geq \lambda_n$ with $\lambda_{2k}=\lambda_{2k+1}$ for $k=1,\ldots,\lfloor (n-1)/2\rfloor$ and $\lambda_{2k-1}>\lambda_{2k}$ for $k=1,\ldots, \lfloor n/2\rfloor$. We denote by $\mathbf v_i$ the eigenvector associated with eigenvalue $\lambda_i$ of $\mathbf L$. 

To generalise the results in Theorem \ref{th:dimensionattraction}, we consider $\mathbf B\in\R^{n\times n}$ with $n-k$ positive eigenvalues and $k$ negative eigenvalues. We denote the  eigenvectors associated with positive eigenvalues by $\mathbf p_1,\ldots,\mathbf p_{n-k}$ and we have $\mathcal W\subset \spann\{ \mathbf p_1,\ldots, \mathbf p_{n-k}\}$  implying $\dim \mathcal W\leq n-k$. In the worst case scenario, we have $\dim \mathcal W=n-k$ which is equal to the dimension of the attraction region of GD and the standard LSGD. However, only a small number of eigenvectors $\mathbf p_j$ for $j\in \{1,\ldots,n-k\}$ usually satisfies $\mathbf L \mathbf p_j\in \spann\{ \mathbf p_j\}$ and hence $\dim \mathcal W$ is much smaller in practice. To see this, note that for any eigenvector $\mathbf p_j$  associated with a positive eigenvalue of $\mathbf B$, we can write $\mathbf p_j=\sum_{i=1}^n \alpha_i \mathbf v_i$ for $\alpha_1,\ldots,\alpha_n\in \R$ where $\mathbf v_i$ are the $n$ eigenvectors of $\mathbf L$. Since $\mathbf L\mathbf p_j=\sum_{i=2}^n \alpha_i \lambda_i \mathbf v_i$, we have $\mathbf L\mathbf p_j\in \spann\{\mathbf p_j\}$ if and only if $\mathbf{p}_j\in \spann\{ \mathbf v_{1}\}$ or $\mathbf{p}_j\in \spann\{ \mathbf v_{2k},\mathbf v_{2k+1}\}$ for some $k\in\{1,\ldots,\lfloor (n-1)/2\rfloor\}$ or, provided $n$ even, $\mathbf{p}_j\in \spann\{ \mathbf v_{n}\}$.

\subsubsection{Degenerate Hessians}
While our approach is very promising for Hessians with both positive and negative eigenvalues, it does not resolve issues of GD or LSGD related to degenerate saddle points where at least one eigenvalue is 0. Let $f(\mathbf x)=\frac{1}{2}\mathbf x^T \mathbf B \mathbf x$ where $\mathbf B\in \R^{n\times n}$ has at least one eigenvalue 0 and let $\mathbf p$ denote an eigenvector associated with eigenvalue 0. 
Then, $\mathbf A_\sigma \mathbf B \mathbf p=\mathbf 0$ for any $\sigma\geq 0$ and hence choosing $\mathbf x^0\in \spann\{\mathbf p\}$ as the starting point for the modified LSGD \eqref{Scheme} will result in $\mathbf x^k=\mathbf x^0$ for all $k\geq 0$ like for GD and LSGD. The investigation of appropriate deterministic perturbations of first-order methods for saddle points where at least one eigenvalue is 0 is subject of future research.

\section{Convergence Rate of the modified LSGD
}\label{Convergence-Rate}

In this section, we discuss the convergence rate of the modified LSGD for iteration-dependent functions $\sigma$ when applied to $\ell$-smooth nonconvex functions $f\colon \R^n\to \R$. Our analysis follows the standard convergence analysis framework. We start with the definitions of the smoothness of the objective function $f$ and a convergence criterion for nonconvex optimization.

\begin{definition}
A differentiable function $f$ is $\ell$-smooth (or $\ell$-gradient Lipschitz), if for all $\mathbf{x},\mathbf{y}\in\R^n$ $f$ satisfies 
$$
f(\mathbf{y}) \leq f(\mathbf{x}) + \nabla f(\mathbf{x})\cdot (\mathbf{y}- \mathbf{x}) + \frac{\ell}{2}\|\mathbf{x}-\mathbf{y}\|^2.
$$
\end{definition}

\begin{definition}
For a differentiable function $f$, we say that $\mathbf{x}$ is an $\epsilon$-first-order stationary point if $\|\nabla f(\mathbf{x})\|\leq \epsilon$.
\end{definition}

\begin{theorem}
Assume that the function $f$ is $\ell$-smooth and let $\sigma$ be a positive, bounded function, i.e., there exists a constant $C>0$ such that $|\sigma(k)|\leq C$ for all $k\in \N$. Then, the modified LSGD with $\sigma$ above,  step size  $\eta={1}/{\ell}$ and termination condition $\|\nabla f(\mathbf{x})\| \leq \epsilon$ converges to an $\epsilon$-first-order stationary point within $$\frac{2(1+4C)^2\ell(f(\mathbf{x}^0)-f^*)}{(1+8C)\epsilon^2}$$ 
iterations, where $f^*$ denotes a local minimum of $f$.
\end{theorem}

\begin{proof}
First, we will establish an estimate for $f(\mathbf{x}^{k+1}) - f(\mathbf{x}^k)$ for all $ k\geq 0$. By the $\ell$-smoothness of $f$ and the LSGD scheme \eqref{Scheme}, we have
\begin{align*}
&f(\mathbf{x}^{k+1}) - f(\mathbf{x}^k) \\ \nonumber
&\leq \langle\nabla f(\mathbf{x}^k), \left(\mathbf{x}^{k+1} - \mathbf{x}^k\right) \rangle + \frac{\ell}{2} \|\mathbf{x}^{k+1} - \mathbf{x}^k\|^2\\ \nonumber
&= \langle \nabla f(\mathbf{x}^k), -\frac{1}{\ell} \mathbf{A}_{\sigma(k)}^{-1}\nabla f(\mathbf{x}^k)\rangle +  \nonumber
\frac{1}{2\ell}\|\mathbf{A}_{\sigma(k)}^{-1}\nabla f(\mathbf{x}^k)\|^2\\ \nonumber
&= \frac{1}{2\ell} \left\|\left(\mathbf{A}_{\sigma(k)}^{-1} -\mathbf{I}\right) \nabla f(\mathbf{x}^k)\right\|^2 - \frac{1}{2\ell} \|\nabla f(\mathbf{x}^k) \|^2\\ \nonumber
&\leq\frac{1}{2\ell} \left\|\mathbf{I}-\mathbf{A}_{\sigma(k)}^{-1}\right\|^2\|\nabla f(\mathbf{x}^k)\|^2 - \frac{1}{2\ell} \|\nabla f(\mathbf{x}^k)\|^2 \nonumber.
\end{align*}
To estimate $\| \mathbf{I}-\mathbf{A}_{\sigma(k)}^{-1}\|$, we note that $\mathbf{A}_{\sigma(k)}^{-1}$ is diagonalizable, i.e., there exists an orthogonal matrix $\mathbf{Q} \in \mathbb{R}^{n\times n}$ and a diagonal matrix $\Lambda$ with diagonal entries $\lambda_j(\mathbf{A}_{\sigma(k)}^{-1}) \in [1, 1+4\sigma(k)]$  such that $\mathbf{A}_{\sigma(k)}^{-1}=\mathbf{Q}^T\boldsymbol{\Lambda} \mathbf{Q}$. We have
\begin{align*}
\|\mathbf{I}-\mathbf{A}_{\sigma(k)}^{-1}\|^2 = \|\mathbf{I}-\boldsymbol{\Lambda}\|^2 \leq\left(1-\frac{1}{1+4\sigma(k)}\right)^2 
\leq \left( \frac{4C}{1+4C}\right)^2.
\end{align*}
Plugging this estimate into the previous estimate yields
$$
f(\mathbf{x}^{k+1}) - f(\mathbf{x}^k) \leq -\frac{1+8C}{2(1+4C)^2\ell} \|\nabla f(\mathbf{x}^k)\|^2.
$$
Based on the above estimate, the function value of the iterates decays by at least $$\frac{1+8C}{2(1+4C)^2\ell} \|\nabla f(\mathbf{x}^k)\|^2$$ in each iteration before an $\epsilon$-first-order stationary point is reached. After $$\frac{2(1+4C)^2\ell(f(\mathbf{x}^0)-f^*)}{(1+8C)\epsilon^2}$$
iterations, the modified LSGD is guaranteed to converge to an $\epsilon$-first-order stationary point with function value $f^*$.

\end{proof}

We note that the above convergence rate for nonconvex optimization is consistent with the gradient descent \cite{Nesterov:1998}, and thus mLSGD converges as fast as GD.

\section{Numerical Examples}\label{Numerical-Results}
In this section, we verify numerically that the modified LSGD does not converge to the unique saddle point in the two-dimensional setting, provided the matrices are not of the special case discussed in Section \ref{sec:extension2d}. We consider the bounded function $\sigma(k)=\frac{k+1}{k+2}$ for the modified LSGD. For both GD and the modified LSGD, we perform an exhaustive search with very fine grid sizes to confirm  our theoretical results empirically. The exhaustive search is computationally expensive, and thus we restrict our numerical examples to the 2-dimensional setting.

\subsection{Example 1}
\begin{figure}
	\centering
	\subfloat[GD]{\includegraphics[width=0.4\textwidth]{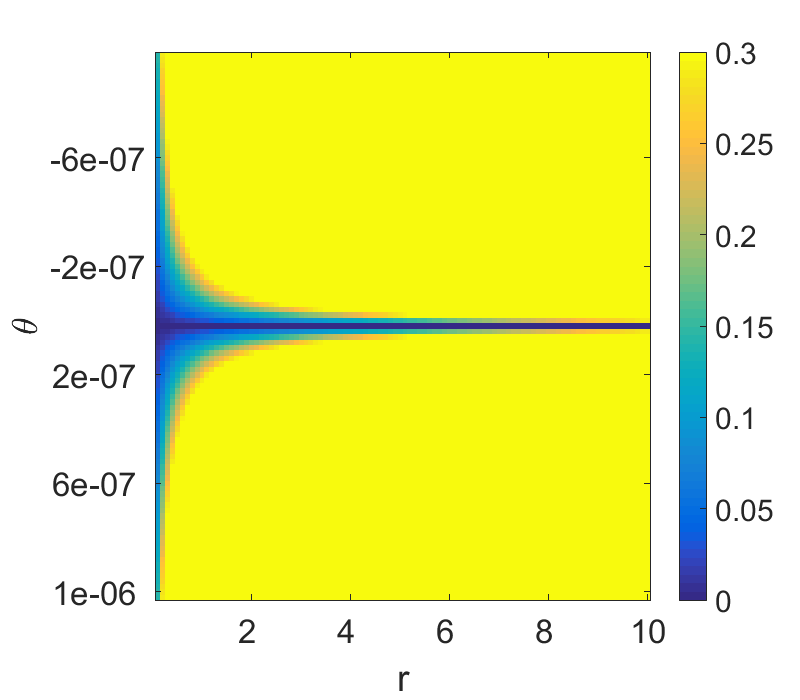}\label{fig:distantgd}}\hspace*{3em}
	\subfloat[Modified LSGD]{\includegraphics[width=0.4\textwidth]{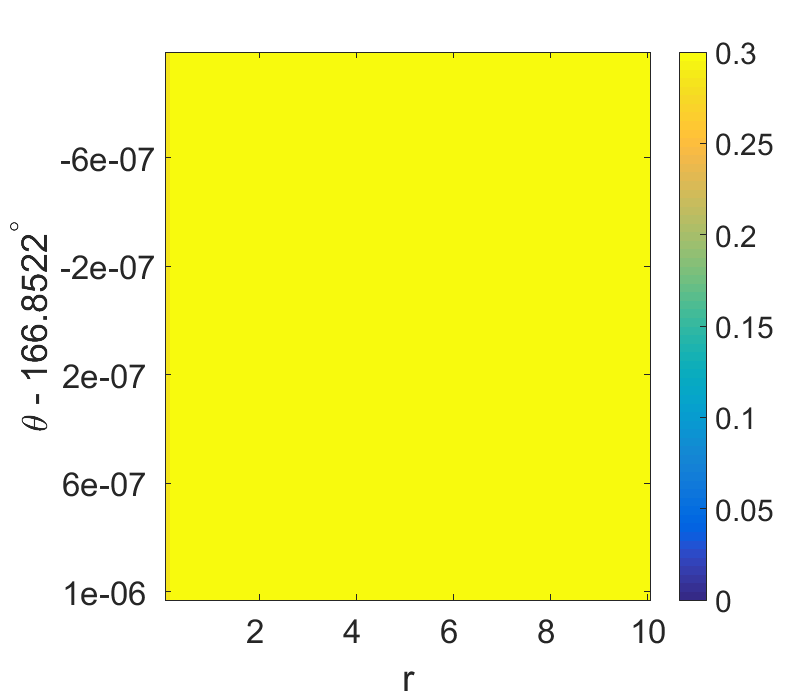}\label{fig:distantlsgd}}
	\caption{Distance field for the saddle point $\mathbf 0$ after  100 iterations for GD  and modified LSGD with learning rate $\eta=0.1$ for the function $f(x_1, x_2) = x_1^2 - x_2^2$ where the coordinates of each pixel denote the starting point and the color shows the distance to the saddle point after 100 iterations.}
	\label{fig:EscapeSaddle}
\end{figure}

We consider the  optimization problem 
\begin{equation}\label{eq:ex1}
\min_{x_1,x_2}f(x_1,x_2):=x_1^2-x_2^2.
\end{equation}
It is easy to see that $[0,0]^T$ is the unique saddle point of $f$. We run $100$ iterations of  GD and the modified LSGD with step size $\eta=0.1$ for solving \eqref{eq:ex1}. 
For GD, the attraction region is given by $\{[x_1, x_2]^T\colon x_1\in \mathbb{R}, x_2 = 0\}$. To demonstrate GD's behavior in terms of its convergence to saddle points, we start GD from any point in the set $\{[x_1, x_2]^T \colon x_1 = r\cos\theta, x_2 = r\sin\theta, r\in[0.1, 10], \theta\in [-1\text{e-6}^\circ, 1\text{e-6}^\circ)\}$, with a grid spacing of $0.1$ and $2\text{e-8}^\circ$ for $r$ and $\theta$, respectively. As shown in Figure \subref*{fig:distantgd}, the distance to the saddle point $[0,0]^T$ is 0 after  100 GD iterations for any starting point with $\theta=0$. For starting points close to $[0,0]^T$, given by small values of $r$ and any $\theta$, the iterates are still very close to the saddle point after $100$ GD iterations with distances less than $0.1$.

For the modified LSGD when applied to solve \eqref{eq:ex1}, the attraction region associated with the saddle point $[0,0]^T$ is of dimension zero, see Theorem \ref{th:dimensionattraction}. To verify this numerically, we consider any starting point in $\{[x_1, x_2]^T | x_1 = r\cos\theta, x_2 = r\sin\theta, r\in[0.1, 1], \theta\in [-180^\circ, 180^\circ)\}$ with a grid spacing of $0.1$ and $1\text{e-6}^\circ$  for $r$ and $\theta$, respectively. We observe that the minimum distance to $[0,0]^T$ is achieved when we start from the point $[r_0 \cos \theta_0,r_0 \sin \theta_0]^T$ for $r_0=0.1$ and $\theta_0=166.8522^\circ$. Then, we perform a finer grid search on the interval $[\theta_0- 1^\circ,\theta_0+1^\circ]$ using  grid spacing $\Delta \theta=2\text{e-8}^\circ$. This two-scale search significantly reduces the computational cost. Figure~\subref*{fig:distantlsgd} shows a similar region as in Figure~\subref*{fig:distantgd}, but with $\theta$ centered at $\theta_0$. If $r=0.1$, the distance to the saddle point is less than $0.3$ but larger than $0.2$, implying that the distance to the saddle point increases by applying $100$ iterations of the modified LSGD.  For any starting point with $r>0.1$ the distance is larger than $0.3$ after $100$ iterations. This illustrates that the iterates do not converge to the saddle point $[0,0]^T$. 

For the 2-dimensional setting, our numerical experiments demonstrate that the modified LSGD does not converge to the saddle point for any starting point provided the conditions in Section \ref{sec:extension2d} are not satisfied. While there exists a region of starting points for GD with a slow escape from the saddle point,  this region of slow escape is significantly smaller for the modified LSGD. These results are consistent with the dimension $\lfloor (n-1)/2\rfloor=0$ of the attraction region for the modified LSGD in Theorem \ref{th:dimensionattraction}. While the analysis is based on the assumption that $\sigma$ is constant at some point, the numerical results indicate that the theoretical results also hold for strictly monotonic, bounded functions $\sigma$, provided $\sigma(k)$ for $k$ large enough is close to being stationary.

\subsection{Example 2}
\begin{figure}
	\centering
	\subfloat[GD]{\includegraphics[width=0.4\textwidth]{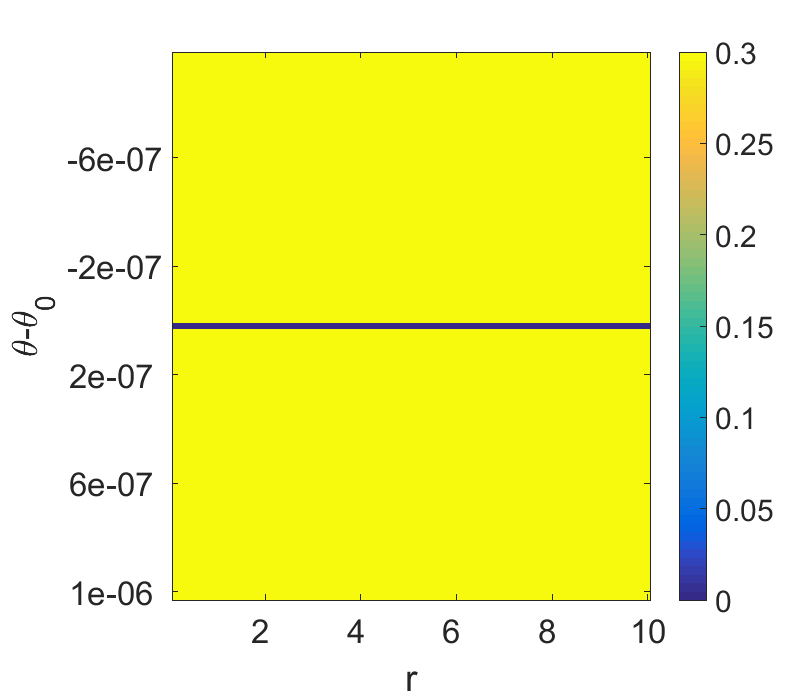}\label{fig:distantgd:ex2}}\hspace*{3em}
	\subfloat[Modified LSGD]{\includegraphics[width=0.4\textwidth]{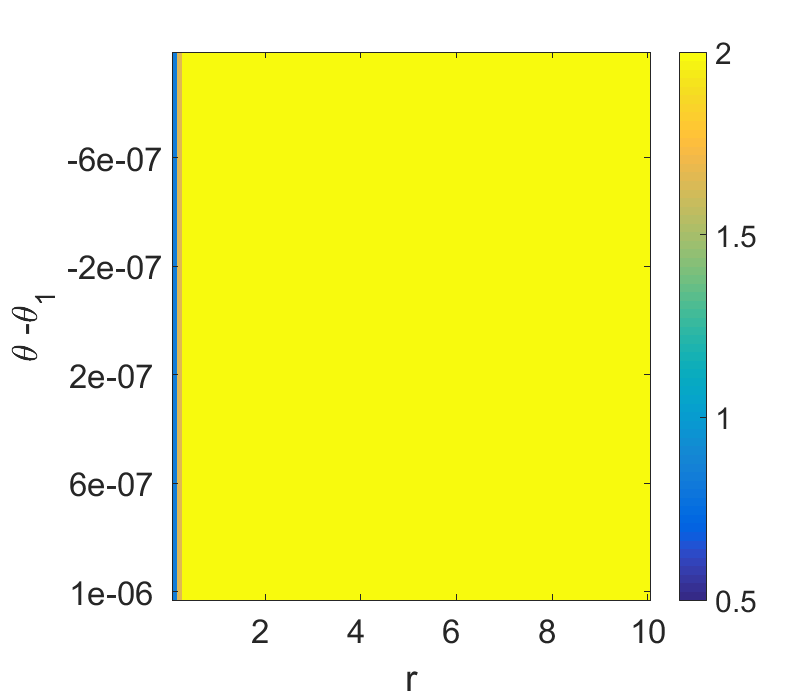}\label{fig:distantlsgd:ex2}}
	\caption{Distance field to the saddle point $\mathbf 0$ after 100 iterations for GD and the modified LSGD with learning rate $\eta=0.1$ for the function $f(x_1, x_2) = x_1^2+6x_1x_2+2x_2^2$ where the coordinates of each pixel denote the starting point and the color shows the distance to the saddle point after 100 iterations ($\theta_0=\arctan\left(\frac{6}{\sqrt{37}-1}\right), \theta_1=-132.635976^\circ$).}
	\label{fig:EscapeSaddle:Ex2}
\end{figure}

To   corroborate  our theoretical findings numerically, we consider a two-dimensional problem where all entries of the coefficient matrix are non-zero. We consider
\begin{equation}\label{eq:ex2}
\min_{x_1,x_2}f(x_1,x_2):=x_1^2+6x_1x_2+2x_2^2
\end{equation}
which satisfies $f(x_1,x_2)=\frac{1}{2} [x_1 x_2] \mathbf B [x_1 x_2]^T$ with
\begin{align*}
    \mathbf B=\begin{bmatrix}2& 6\\ 6&4\end{bmatrix}.
\end{align*}
We apply GD with step size $\eta$ and starting from $[x_1^0,x_2^0]^T$ for solving \eqref{eq:ex2}, resulting in the  iterations
\begin{align}\label{GD:ex2}
\begin{bmatrix}
x_1^{k+1}\\ x_2^{k+1}
\end{bmatrix}=\begin{bmatrix}
x_1^{k}\\ x_2^{k}
\end{bmatrix} - \eta\begin{bmatrix}
2x_1^k+6x_2^k\\ 6x_1^k+4x_2^k 
\end{bmatrix} = \begin{bmatrix}
x_1^{k}\\ x_2^{k}
\end{bmatrix} + \eta\begin{bmatrix}
-2&-6\\ -6&-4
\end{bmatrix} \begin{bmatrix}
x_1^{k}\\ x_2^{k}
\end{bmatrix}.
\end{align}
The eigenvalues of the coefficient matrix  $\mathbf B$ are $\lambda_1=\sqrt{37}+3$ and $\lambda_2=-\sqrt{37}+3$, and the associated eigenvectors are 
\begin{align*}
\mathbf{v}_1=\begin{bmatrix}\frac{\sqrt{37}-1}{6}\\ 1\end{bmatrix} \quad \text{and} \quad \mathbf{v}_2=\begin{bmatrix}
\frac{-\sqrt{37}-1}{6}\\ 1\end{bmatrix},
\end{align*}
respectively. 
If $[x_1^0,x_2^0]^T$ is in $\spann\{\mathbf{v}_1\}$, GD  converges to the saddle point $[0,0]^T$. As shown in Figure \subref*{fig:distantgd:ex2}, starting from any point in 
$\spann\{[\cos \theta,\sin\theta]^T\}$ 
with 
$${\theta=\arctan\left(\frac{6}{\sqrt{37}-1}\right)},$$
$[x_1^k,x_2^k]^T$ converges to the unique saddle point after 100 iteration. 
To corroborate our theoretical result that the modified LSGD does not converge to the saddle point in two dimensions, we perform a two-scale exhaustive search. First, we search over the initial point set $\{[x_1, x_2]^T | x_1 = r\cos\theta, x_2 = r\sin\theta, r\in[0.1, 1], \theta\in [-180^\circ, 180^\circ)\}$ with  grid spacing of $0.1$ and $1\text{e-6}^\circ$ for $r$ and $\theta$, respectively. We observe that the minimum distance to $[0,0]^T$ is achieved when we start from the point $[r_0 \cos \theta_0,r_0 \sin \theta_0]^T$ for $r_0=0.1$ and $\theta_0=-132.635976^\circ$. Then, we perform a finer grid search on the interval $[\theta_0- 1^\circ,\theta_0+1^\circ]$ using the grid spacing $\Delta \theta=2\text{e-8}^\circ$. Figure~\subref*{fig:distantlsgd:ex2} shows a similar region as in Figure \subref*{fig:distantgd:ex2}, but with $\theta$ centered at $\theta_0$. After $100$ LSGD iterations, the iterates do not converge to the saddle point $[0,0]^T$, and we note that the minimum distance to the saddle point $[0,0]^T$ is $0.83$.

\section{Concluding Remarks}\label{Conclusion}
In this paper, we presented a simple modification of the Laplacian smoothing gradient descent (LSGD) to avoid saddle points. We showed that the modified LSGD can efficiently avoid saddle points both theoretically and empirically. In particular, we proved that the modified LSGD can significantly reduce the dimension of GD's attraction region  for a class of quadratic objective functions. Nevertheless, our current modified LSGD does not reduce the attraction region when applied to minimize some objective functions, e.g., $f(x_1, x_2) = x_1x_2$. It is interesting to extend the idea of modified LSGD to avoid saddle points for general objective functions in the future.

To the best of our knowledge, our algorithm is the first deterministic gradient-based algorithm for avoiding saddle points that leverages only first-order information without any stochastic perturbation or noise. Our approach differs from existing perturbed or noisy gradient-based approaches for avoiding saddle points. It is of great interest to investigate the efficacy of a combination of these approaches in the future. A possible avenue is to integrate Laplacian smoothing with perturbed/noisy gradient descent to escape and circumvent saddle points more efficiently.

\section*{Acknowledgments}
This material is based on research sponsored by the National Science Foundation under grant numbers DMS-1924935, DMS-1952339 and DMS-1554564 (STROBE), the Air Force Research Laboratory under grant numbers FA9550-18-0167 and MURI FA9550-18-1-0502, the Office of Naval Research under the grant number N00014-18-1-2527, and the Department of Energy under the grant number DE-SC0021142. LMK acknowledges support from the UK Engineering and Physical Sciences Research Council (EPSRC) grant EP/L016516/1,  the German National Academic Foundation (Studienstiftung des Deutschen Volkes), the European Union Horizon 2020 research and innovation programmes under the Marie Skłodowska-Curie grant agreement No. 777826 (NoMADS) and No. 691070 (CHiPS), the Cantab Capital Institute for the Mathematics of Information and Magdalene College, Cambridge (Nevile Research Fellowship).

\bibliographystyle{siam.bst}
\bibliography{references.bib}

\end{document}